\documentclass{article}

\usepackage{arxiv}
\usepackage[numbers,sort&compress]{natbib}
\usepackage[utf8]{inputenc}
\usepackage[T1]{fontenc}
\usepackage[htt]{hyphenat}   % allow line breaks inside \texttt paths
\usepackage{amsmath}
\usepackage{amssymb}
\usepackage{microtype}
\usepackage{graphicx}
\usepackage{booktabs}
\usepackage{xcolor}
\usepackage{multirow}
\usepackage{tabularx}
\usepackage{listings}
\usepackage{framed}
\usepackage{algorithm}
\usepackage{algpseudocode}
\usepackage{tikz}
\usetikzlibrary{positioning,arrows.meta,calc,decorations.pathreplacing}

\definecolor{adalpink}{HTML}{FF5898}   % the brand pink itself
\definecolor{pinkA}{HTML}{FDEAF2}      % lightest tint  — large fills
\definecolor{pinkB}{HTML}{F9C6DA}      % light tint
\definecolor{pinkC}{HTML}{EE8CB4}      % mid
\definecolor{pinkDeep}{HTML}{B02A62}   % deep — white text sits on this
\definecolor{pinkInk}{HTML}{8E2350}    % text/rules in the pink family
\definecolor{adalgreen}{HTML}{9BE564}
\definecolor{greenInk}{HTML}{4E7A2A}

\newcolumntype{L}[1]{>{\raggedright\arraybackslash}p{#1\textwidth}}
\newcolumntype{C}[1]{>{\centering\arraybackslash}p{#1\textwidth}}

\newcommand{\minisection}[1]{\vspace{1mm}\noindent{\textbf{#1}~}}
\newcommand{\atskill}{\texttt{@skills}}
\newcommand{\todo}[1]{}

\usepackage[breaklinks,colorlinks,linkcolor=blue!60!black,citecolor=blue!60!black,urlcolor=blue!60!black]{hyperref}
\usepackage[capitalize]{cleveref}

\crefname{section}{Sec.}{Secs.}
\Crefname{section}{Section}{Sections}
\Crefname{table}{Table}{Tables}
\crefname{table}{Tab.}{Tabs.}

\begin{document}

\title{\texttt{@skills}: Attention Is All You Have\\[0.5em]
  {\large\normalfont An open agent skills protocol, built for the future of knowledge sharing}}

\author{Li Yin\textsuperscript{1} \quad Zhi Li\textsuperscript{1} \quad Zhan Shi\textsuperscript{1} \quad Haoran Zhang\textsuperscript{1} \quad Haebin Seong\textsuperscript{1} \quad Zhangyang (Atlas) Wang\textsuperscript{2}\thanks{The open-source protocol---specification, agent integration guide, and example workflows---is at \url{https://github.com/SylphAI-Inc/atskills}. Correspondence via the repository's issue tracker.}\\
  \textsuperscript{1}SylphAI \qquad \textsuperscript{2}The University of Texas at Austin}

\date{}

\renewcommand{\shorttitle}{\texttt{@skills}: Attention Is All You Have}

\maketitle

\begin{abstract}
Agent skills package procedural knowledge as \texttt{SKILL.md} files: 56{,}804 are published today, and teams write many more privately. The dominant way to deliver one is to install it, after which its description sits permanently in the system prompt, competing for fewer than a hundred reliable slots in which the model may or may not match it to a request. So the long tail has no path to use, and a team's own playbooks compete with everything else for the same scarce room. Our observation is that installing bundles three separable functions---content, persistence, and auto-triggering---and only the last needs the prompt. We propose \atskill{}, an open protocol that separates them. A path addresses any skill, any subtree, or a whole collection, and reading it is using it, so nothing installs and nothing becomes resident; \texttt{:save} vendors a copy at that same path into the project's git-tracked tree, to adapt and own it; \texttt{:install} adds one \texttt{.gitignore}-style line, which is the only thing that costs prompt residency. A directory is a menu, so a ``bundle'' is just a directory and all-or-nothing delivery does not exist. There is no manifest, lockfile, or registration, because a file tree needs none, and \texttt{SKILL.md} is unchanged. The protocol is additive, ships as an installable package with an open specification (\url{https://github.com/SylphAI-Inc/atskills}), and turns any agent that can read files and run commands into a client from a single instruction file. The protocol is implemented in the AdaL CLI (\url{https://adalagent.ai}). Because a file tree addresses skills well but cannot find them, the protocol also pairs with a free agent skills hub (\url{https://atskills.one}) confined to the jobs paths cannot do: search and ranking across the whole public corpus, hosting for skills that have no repository, private and team collections, and one-screen authoring for the non-developers whose procedural knowledge skills capture best. The hub is a service and never a requirement---\texttt{gh:} and local paths resolve with no hub involvement, and GitHub-hosted skills keep their \texttt{gh:} identity even when the hub indexes them. Install less, use more.
\end{abstract}


% ── Main body (short by design; the evidence loads on reference) ──
\section{Introduction}
\label{sec:intro}

Coding agents increasingly extend themselves with \emph{skills}: packages of procedural knowledge, such as how to produce a changelog, deploy a service, or typeset a paper, written as Markdown instructions in a standardized \texttt{SKILL.md} file. Our July 2026 crawl of the public ecosystem finds 56{,}804 such skills across 1{,}133 GitHub repositories, indexed by community registries such as skills.sh \citep{vercel2025skills}. The official definition says what a skill \emph{is}:

\begin{quote}
\itshape A skill is a directory containing a \texttt{SKILL.md} file: organized folders of instructions, scripts, and resources that give agents additional capabilities.
\normalfont\hfill ---the official definition \citep{anthropic2025skills,anthropic2025skillseng}
\end{quote}

What the published corpus shows is what skills have become \emph{for}, and this paper's argument rests on that shift, so we extend the definition (evidence in \cref{app:landscape}):

\begin{quote}
\itshape A skill is procedural knowledge for an AI agent, either the world's expertise or a team's own way of doing things, and it serves three roles: \emph{domain knowledge}, expertise distilled into instructions; \emph{provider integration}, by which a service teaches any agent to drive its product; and \emph{team workflows}, the customized procedures a team adapts to its own way of working. A model knows what the world knows; a skill adds what it doesn't.
\end{quote}

Skills spread because they are simpler than what came before. The Model Context Protocol \citep{anthropic2024mcp} standardized agent-to-tool connectivity but is heavyweight on both sides, since the provider maintains a server and the user pays for resident tool schemas before any work begins, whereas a skill is a folder of text the agent simply reads \citep{willison2025skills}. Yet installed skills kept MCP's delivery model in miniature: something is still installed per machine, and just as MCP loads every server's tool definitions at session start, an installed skill puts its description into the system prompt in every session, relevant or not. The artifact got radically simpler; the \emph{residency} did not.

That inheritance is the subject of this paper. Installation is what an agent skill's description buys with permanent space in the prompt, and the only thing it buys is \emph{auto-triggering}---firing without being asked. Auto-triggering is also scarce, with a reliable capacity well under a hundred slots per agent, so 56{,}804 published skills compete for fewer than a hundred (\cref{sec:problem}). Everything else installation carries---fetching the content, and keeping it for next time---needs no prompt space at all.

We propose \atskill{}, which separates those three functions and lets each be decided on its own (\cref{fig:tiers,fig:split}). A path addresses any skill, any subtree, or a whole collection and is read at the point of use; \texttt{:save} keeps a copy in the project's git-tracked tree; \texttt{:install} adds one line to a \texttt{.gitignore}-style file, the only thing that costs residency (\cref{sec:protocol}). The management model is nothing but the file tree, so there is no manifest, lockfile, or registration, and \texttt{SKILL.md} is untouched.

\minisection{Terminology.} A \emph{skill} is a directory, not just a Markdown file: \texttt{SKILL.md} holds frontmatter with a name and a one-line description followed by instructions, alongside optional scripts and reference files that load only when needed. A \emph{plugin} is an installable bundle and a \emph{marketplace} is a registry of bundles, though the two terms have no shared definition across agents (\cref{app:background}); both inherit the install lifecycle. We call text \emph{resident} when it occupies the prompt on every message of a session. The body of this paper carries the argument; the evidence, mechanics, and full protocol are in the appendices, loaded on reference.

\begin{figure}[t]
\centering
\begin{tikzpicture}[line join=round,
    lft/.style={anchor=east, align=right, text=black!60, font=\scriptsize},
    rgt/.style={anchor=west, align=left,  text=black!60, font=\scriptsize}]
  \colorlet{tierone}{pinkA}
  \colorlet{tiertwo}{pinkB}
  \colorlet{tierthree}{pinkDeep}

  % ── callout: the whole ecosystem is crammed into the apex ──
  \node[align=center, font=\scriptsize\scshape, text=black!65] at (0,4.40)
    {the existing ecosystem offers only this tier};
  \draw[-{Stealth[length=4pt]}, black!40] (0,4.22) -- (0,3.84);

  % ── column headers ──
  \node[lft, text=black!50, font=\scriptsize\scshape] at (-4.5,4.10) {cost $\cdot$ how it fires};
  \node[rgt, text=black!50, font=\scriptsize\scshape] at ( 4.5,4.10) {what belongs here};

  % ── Tier 3 (apex) ──
  \draw[fill=tierthree, draw=white, line width=1.2pt]
    (-1.90,3.70) -- (1.90,3.70) -- (2.55,2.65) -- (-2.55,2.65) -- cycle;
  \node[text=white, font=\footnotesize\bfseries] at (0,3.36) {Tier 3 --- Installed};
  \node[text=white!85, font=\scriptsize] at (0,3.03)
    {\texttt{:install} $\cdot$ under 10};

  % ── Tier 2 (middle) ──
  \draw[fill=tiertwo, draw=white, line width=1.2pt]
    (-2.55,2.65) -- (2.55,2.65) -- (3.40,1.25) -- (-3.40,1.25) -- cycle;
  \node[font=\footnotesize\bfseries] at (0,2.28) {Tier 2 --- Saved / Project workflows};
  \node[font=\scriptsize, text=black!70] at (0,1.95)
    {\texttt{:save} $\cdot$ 10--30 $\cdot$ git-tracked in \texttt{.atskills/}};

  % ── Tier 1 (base) ──
  \draw[fill=tierone, draw=white, line width=1.2pt]
    (-3.40,1.25) -- (3.40,1.25) -- (4.20,-0.25) -- (-4.20,-0.25) -- cycle;
  \node[font=\footnotesize\bfseries] at (0,0.82) {Tier 1 --- Reference};
  \node[font=\scriptsize, text=black!70] at (0,0.49)
    {\texttt{@skills:<path>} $\cdot$ 56{,}804+ $\cdot$ nothing stored};

  % ── left: what it costs on every message ──
  \node[lft] at (-4.5,3.20)
    {50--280 tok/skill, every message\\ \textcolor{black!45}{implicit --- fires unasked}};
  \node[lft] at (-4.5,1.98)
    {0 tokens\\ \textcolor{black!45}{explicit --- autocompleted}};
  \node[lft] at (-4.5,0.62)
    {0 tokens\\ \textcolor{black!45}{explicit --- you name the path}};

  % ── right: the scenarios each tier is for ──
  \node[rgt] at (4.5,3.20) {formatter conventions,\\ security guardrails};
  \node[rgt] at (4.5,1.98) {the deploy runbook,\\ review checklist, migrations};
  \node[rgt] at (4.5,0.62) {trying one out, comparing,\\ one-off needs --- the long tail};
\end{tikzpicture}
\caption{The three delivery tiers, sized by how many skills each is for. A skill belongs in the \emph{cheapest tier that meets its activation need}, and only deliberate promotion moves it up---so nothing persists by accident. Tiers 1 and 2 differ solely in where the bytes live: nowhere, or vendored in the project's git-tracked tree, where the \texttt{@} autocomplete indexes them in the tooling rather than in the prompt. Neither costs a resident token, and both fire explicitly, by name, at the point of use. Tier 3 alone pays residency, and buys exactly one thing with it---firing without being asked---for its frontmatter alone. The existing ecosystem offers only that apex, so all 56{,}804 skills compete for an attention budget argued in \cref{app:failures} to be well under a hundred slots per agent.}
\label{fig:tiers}
\end{figure}
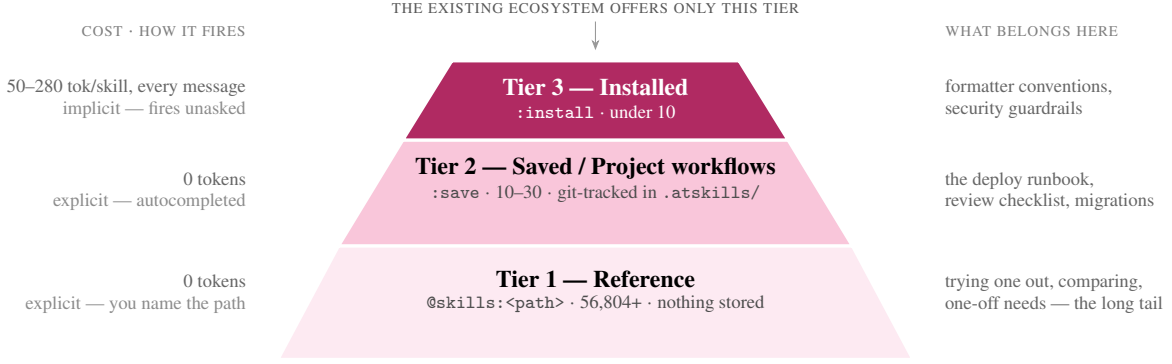
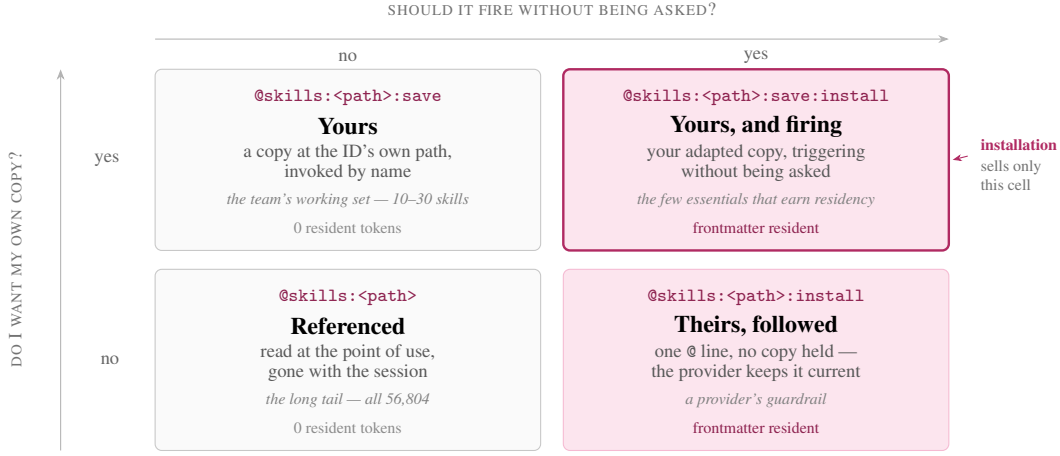
\begin{figure}[t]
\centering
\begin{tikzpicture}[line join=round,
    cell/.style={rounded corners=2.5pt},
    ttl/.style={font=\footnotesize\bfseries, align=center},
    sfx/.style={font=\scriptsize\ttfamily, align=center, text=pinkInk},
    dsc/.style={font=\scriptsize, align=center, text=black!65},
    use/.style={font=\tiny\itshape, align=center, text=black!50},
    cst/.style={font=\tiny, align=center},
    ax/.style={font=\scriptsize\scshape, text=black!55},
    lvl/.style={font=\scriptsize, text=black!60}]

  % ── the two axes ──
  \node[ax] at (7.45,6.15) {should it fire without being asked?};
  \draw[-{Stealth[length=4pt]}, black!30] (2.2,5.75) -- (12.7,5.75);
  \node[lvl] at (4.75,5.52) {no};
  \node[lvl] at (10.15,5.52) {yes};

  \node[ax, rotate=90] at (0.35,2.83) {do I want my own copy?};
  \draw[-{Stealth[length=4pt]}, black!30] (0.95,0.3) -- (0.95,5.35);
  \node[lvl, anchor=east] at (1.85,1.50) {no};
  \node[lvl, anchor=east] at (1.85,4.15) {yes};

  % ══ top-left: :save — owned, explicit ══
  \fill[cell, black!2] (2.2,2.95) rectangle (7.3,5.35);
  \draw[cell, black!22] (2.2,2.95) rectangle (7.3,5.35);
  \node[sfx] at (4.75,4.98) {@skills:<path>:save};
  \node[ttl] at (4.75,4.60) {Yours};
  \node[dsc] at (4.75,4.24) {a copy at the ID's own path,};
  \node[dsc] at (4.75,3.98) {invoked by name};
  \node[use] at (4.75,3.62) {the team's working set --- 10--30 skills};
  \node[cst, text=black!45] at (4.75,3.25) {0 resident tokens};

  % ══ top-right: :save:install — owned, automatic ══
  \fill[cell, pinkB!45] (7.6,2.95) rectangle (12.7,5.35);
  \draw[cell, pinkC!60] (7.6,2.95) rectangle (12.7,5.35);
  \node[sfx] at (10.15,4.98) {@skills:<path>:save:install};
  \node[ttl] at (10.15,4.60) {Yours, and firing};
  \node[dsc] at (10.15,4.24) {your adapted copy, triggering};
  \node[dsc] at (10.15,3.98) {without being asked};
  \node[use] at (10.15,3.62) {the few essentials that earn residency};
  \node[cst, text=pinkInk] at (10.15,3.25) {frontmatter resident};

  % ══ bottom-left: no suffix — nothing kept ══
  \fill[cell, black!2] (2.2,0.30) rectangle (7.3,2.70);
  \draw[cell, black!22] (2.2,0.30) rectangle (7.3,2.70);
  \node[sfx] at (4.75,2.33) {@skills:<path>};
  \node[ttl] at (4.75,1.95) {Referenced};
  \node[dsc] at (4.75,1.59) {read at the point of use,};
  \node[dsc] at (4.75,1.33) {gone with the session};
  \node[use] at (4.75,0.97) {the long tail --- all 56{,}804};
  \node[cst, text=black!45] at (4.75,0.60) {0 resident tokens};

  % ══ bottom-right: :install — followed, automatic ══
  \fill[cell, pinkB!45] (7.6,0.30) rectangle (12.7,2.70);
  \draw[cell, pinkC!60] (7.6,0.30) rectangle (12.7,2.70);
  \node[sfx] at (10.15,2.33) {@skills:<path>:install};
  \node[ttl] at (10.15,1.95) {Theirs, followed};
  \node[dsc] at (10.15,1.59) {one \texttt{@} line, no copy held ---};
  \node[dsc] at (10.15,1.33) {the provider keeps it current};
  \node[use] at (10.15,0.97) {a provider's guardrail};
  \node[cst, text=pinkInk] at (10.15,0.60) {frontmatter resident};

  % ── what installation offers: one cell, taken or not ──
  \draw[pinkDeep, line width=0.9pt, rounded corners=2.5pt]
    (7.6,2.95) rectangle (12.7,5.35);
  \node[font=\tiny\bfseries, text=pinkDeep, anchor=west] at (13.0,4.35)
    {installation};
  \node[font=\tiny, text=black!55, anchor=west, align=left] at (13.0,3.95)
    {sells only\\ this cell};
  \draw[-{Stealth[length=3.5pt]}, pinkDeep] (12.95,4.20) -- (12.75,4.15);
\end{tikzpicture}
\caption{The contribution in one picture: installation bundles three functions, and the protocol separates them, so that what was a single switch becomes two independent decisions. \textbf{Left to right} is whether the skill should fire without being asked, which is the only thing that requires prompt residency; \textbf{bottom to top} is whether the project keeps a copy of its own. The two suffixes set these separately---\texttt{:install} writes one \texttt{.gitignore}-style line, \texttt{:save} vendors a copy at the ID's own path---and neither implies the other, so all four cells are reachable and each is right somewhere. Installing offers only the upper right, and offers it as the sole way to obtain content at all, which is why every one of 56{,}804 published skills has had to bid for an attention budget of fewer than a hundred slots in order to be used even once. Reading a skill costs nothing that persists, keeping one costs a folder in the repository, and only firing unprompted costs the prompt.}
\label{fig:split}
\end{figure}

\section{The Problem: Attention Is a Limited Budget, and Installation Spends It}
\label{sec:problem}

Installing a skill spends a slot in the resident prompt, and that budget is physically small: our evidence bounds it conservatively at fewer than a hundred reliable auto-trigger slots per agent. Yet installation is the only \emph{persistent} way any of 56{,}804 published skills can reach an agent. A distribution channel is supposed to scale with the corpus, whereas an attention budget cannot, because it is capped by the model's ability to attend. Everything in this section follows from that mismatch:

\begin{center}
\emph{56{,}804 indexed skills compete for fewer than 100 reliable auto-trigger slots per agent.}
\end{center}

The rest of this section is the evidence, at three levels: the cap is real (Level 1), the parties adapt to it in ways that consume more of it (Level 2), and because install is also the only \emph{lifecycle}, the ecosystem inherits structural damage that no amount of budget would fix (Level 3). Nothing here indicts auto-triggering itself, which is genuinely valuable and which the tier model keeps for the instructions that need it; the problem is that one mechanism carries everything. \Cref{app:failures} develops all six mechanisms in full.

\minisection{Level 1 --- the cap is real.} Three properties of resident context bound what it can carry. \emph{Distance decay}: a resident description sits at a fixed place, the top of the context, while every turn, file read, and tool result lands between it and the current request; adherence to the system message measurably decays over turns \citep{liu2024lost,qin2024sysbench,laban2025lost}. \emph{The standing tax}: each installed description is paid on every message, relevant or not---50--280 tokens per skill in our measurements---and input length alone degrades reasoning well below the nominal context limit \citep{levy2024same,hsieh2024ruler}. \emph{Dilution}: each added description competes with the rest, and instruction-following degrades systematically as concurrent instructions accumulate \citep{shi2023distracted,jaroslawicz2025instructions}. Bundling amplifies all three, since the dominant harness offers no per-skill control inside a plugin \citep{claudecode2026plugins}. Together they yield the bound above (\cref{sec:failure-structural}).

\minisection{Level 2 --- the adaptations make it worse.} A scarce channel is one every party works around, and each workaround consumes more of what is scarce.

\emph{Authors bid for slots.} A skill fires only if the model matches the user's message against its one-line description, and a miss is silent---the user never learns the skill was there. So authors stop writing descriptions and start writing bids, padding that one line with trigger and skip conditions: \emph{read this BEFORE opening the file; don't skip because it looks like a one-liner; trigger whenever the prompt mentions \ldots; skip only when \ldots}. Stripe's flagship skill carries roughly 150 words of them. Rational individually, ruinous collectively: such descriptions cost roughly $20\times$ the tokens of a quiet one, so every bid deepens the tax and the dilution just measured (\cref{sec:failure-trigger}).

\emph{Users install less than they could.} A permanent cost is worth paying only for a skill used often, genuinely better fired unprompted, and known to trigger reliably. Trying something out, comparing alternatives, and one-off needs all fail at least one bar while paying full price---permanence, plus management in every agent and on every machine. The rational move is not to bother, which is how most of the corpus goes unexplored.

\emph{What is installed gets forgotten.} Installing is the easy part. The skill then sits in a hidden directory and is supposed to fire on its own, so the user stops tracking it. When it fails to fire, nothing says so---and they cannot invoke it by name either, because they no longer remember it is there.

\emph{Teams skip the channel entirely}, pasting their workflows into a monolithic \texttt{AGENTS.md}: always loaded, zero setup, no trigger lottery. The crudest mechanism wins on the only axis users feel---simplicity.

\minisection{Level 3 --- the damage outlasts the budget.} Suppose the slots were free tomorrow. Three failures would remain, because they come from install being the only \emph{lifecycle} rather than from the cap. First, everything lands in one flat bin: a skill grabbed once for a one-off task sits forever beside the playbooks a team runs daily, indistinguishable from them, so the procedures that matter most have no dedicated home. Second, no shared vocabulary exists to build one from---the layer above the skill, the ``plugin,'' \emph{contains} skills in one agent, sits \emph{beside} them as a sibling category in another, and is \emph{one of seven} peer extension types in a third \citep{claudecode2026pluginsref,openai2026codexplugins,cursor2026customize}---so each agent ships its own hierarchy and its own inventory screen over what are, on disk, the same Markdown directories. Third, none of that machinery serves private work: it grew out of publishing, aimed at distributing skills to strangers, and a team whose skills will never leave the company is served by none of it (\cref{app:background}).

The long tail therefore has no delivery path, authors see no measurable usage and stop publishing, and users conclude the ecosystem is thin. The install lifecycle, not the file format, is what strangles it.

\section{The \texorpdfstring{\atskill{}}{@skills} Protocol}
\label{sec:protocol}
\label{sec:tiers}

\minisection{The design principle: three tiers.} Every skill currently pays the same maximal cost, permanent residency, for the same scarce benefit, a chance at auto-triggering, whether it is used once a year or on every message. The fix is to let each skill sit in the \emph{cheapest tier that meets its activation need}, which is what separating content, persistence, and triggering makes possible: three tiers (\cref{fig:tiers}), of which the ecosystem offers only the most expensive. \Cref{app:tiers} develops their storage and why the protocol defines no user-level install.

\minisection{What it is, and where to get it.} The protocol is four files and one surface: a reference command (\texttt{@skills:<path>}), a project folder (\texttt{.atskills/}), a trigger file (\texttt{.autotrigger}), and a provenance stamp (\texttt{.source}, two lines recording where a saved copy came from and at which revision), plus a management surface (\texttt{/skills}) that every client should ship. The unit of content is the unmodified \texttt{SKILL.md} directory, so this is a delivery layer rather than a fork of the format. All of it ships as one installable package, \texttt{npm i atskills}, holding the normative specification, a TypeScript core, a reference CLI, an agent instruction file, and a runnable demo \citep{agentatskills}; none of what follows is a proposal, since it is implemented, tested, and in production in AdaL \citep{adal2026}. \Cref{app:protocol} gives the full specification and \cref{app:algorithms} the algorithms.

\minisection{Syntax.} A skill is referenced inline, by the user or by the agent itself. The path is the identity, and the two suffixes are orthogonal and combinable:

\begin{lstlisting}
@skills:deploy                    the project's own -- .atskills/deploy/
@skills:hub:sylphai/glowmotion    one skill, from the hub
@skills:gh:acme/skills/deploy     one skill, straight from GitHub
@skills:gh:stripe/agent-toolkit   a directory -> a menu, one line per skill
@skills:<path>:save               copy into the project -- to adapt it
@skills:<path>:install            one line in .autotrigger -- fires on its own
\end{lstlisting}

This reuses the most widely adopted gesture in coding agents, since \texttt{@} is already how users add files and directories in a dozen of them \citep{cursor2025atsymbols,github2024copilot,cline2026docs,sourcegraph2026docs}, and skills simply become one more addressable resource. A path with no \texttt{SKILL.md} is a directory rather than a failure, and it loads as a \emph{menu}: one line per skill beneath it, each line itself a valid path. Browsing and using are therefore the same gesture, a collection can be taken subtree by subtree, and granularity belongs to the reader instead of whoever packaged the bundle.

\minisection{Composition.} Several references load in one message, five of them in \cref{fig:hero}, each injected at its own point of use, and they chain across a workflow's steps. Because every reference is explicit, co-firing $N$ skills is deterministic, whereas under installation it is a lottery in which $N$ descriptions must each win a probabilistic match at once, with the odds compounding against exactly the multi-skill workflows real work is made of (\cref{sec:failure-dilution}). Sources also mix freely within a message, so a workflow is no longer confined to what one publisher shipped together. This is the capability install-only delivery cannot reproduce at any budget, and the nearest non-installing tool in the ecosystem does not have it either, resolving exactly one skill per invocation and refusing any reference that matches more than one \citep{vercel2025skills}. Composition has to be a property of the address, not of the fetch.

\minisection{Resolution: the prefix decides.} A reference states where it comes from, so resolution never guesses: a bare path is the project's own and never reaches the network, while \texttt{hub:} and \texttt{gh:} name the cloud. Both cloud forms still resolve local-first, because a saved copy sits at its ID's own path and therefore answers its own address, which is vendoring as in Go's \texttt{vendor/} directory, and which is what makes the team's adaptation the meaning of that address inside the project. Otherwise the content comes through one machine-wide validating cache (\texttt{\textasciitilde/.cache/atskills}, required to sit outside any \texttt{.atskills/} directory) that behaves like a browser: every use asks the source whether anything changed, in a single revision probe rather than a re-download, so that unchanged serves instantly, changed fetches fresh, and offline serves the cached copy marked stale (\cref{alg:resolve}). No manifest is ever consulted, because none exists, and cache entries are always safe to delete.

\minisection{Ownership: theirs, or yours.} Existing install systems leave a skill in an ambiguous middle state, installed but stale, enabled but of unknown origin, or pending an update that may never be applied. The protocol admits two relationships and eliminates the third by construction. A skill is \emph{theirs} when one \texttt{@} line follows the provider's copy without holding one locally, which is the appropriate arrangement precisely because the provider is documenting a service they continue to change. It becomes \emph{yours} when \texttt{:save} copies it to its ID's own path and detaches it at that moment, so the copy is thereafter an ordinary file of the project rather than a subscription. This is also why there is no update lifecycle and no version pinning: frozen text against a moving service offers only the appearance of safety. Provenance is two lines in \texttt{.source}, written once and never consulted during resolution, and a user who wants a saved copy refreshed simply saves again, which replaces an unedited copy and refuses an edited one rather than overwriting work (\cref{alg:save}).

\minisection{Auto-triggering: one line in one file.} \texttt{.autotrigger} follows \texttt{.gitignore} conventions, one entry per line with \texttt{\#} for comments, where a plain line matches the project's own skills (so \texttt{team-flows/} triggers everything beneath it), an \texttt{@} line follows a cloud skill, and a trailing \texttt{/} takes a whole directory. Installing therefore consists of adding a line and uninstalling of removing one, and nothing further from the classic install lifecycle remains because nothing further proved necessary. What a line costs is the skill's frontmatter alone, some 50--100 tokens held from session start, the body loading only when the skill fires (\cref{alg:residency}). The suffixes, the \texttt{/skills} checkbox tree, and hand edits all write these same lines, and \texttt{/skills} adds a view no agent currently ships: the exact resident text the model will receive, with its token count.

\minisection{What the separation buys.} Because the two suffixes are independent, a user answers two questions instead of one, whether to own a copy and whether the skill should fire unprompted, where installation poses a single bundled question and answers it with content, persistence, and triggering together. All four combinations are then reachable and each is right somewhere (\cref{fig:split}). Two further properties follow. Reversal is symmetric, since uninstalling deletes the line that installing added and nothing else was created to undo, and everything stays reviewable, since what fires unprompted is a single file of one-line diffs while what the team owns is a folder in the repository. Together these turn the cost of a skill from a default into a decision, which is what allows 56{,}804 skills to be reachable while fewer than ten are resident.

\minisection{Adoption: one file, or one CLI.} The protocol asks nothing of the agent or its vendor. \texttt{SKILLS.md}, a single instruction file, turns any agent that can read files, run shell commands, and fetch URLs into a full client, resolution and cache and save and trigger rules included; agents that prefer to shell out call the reference CLI (\texttt{atskills get / save / triggers / prompt}) and inherit the same behavior without implementing it. Either way a resolved skill is just files on disk, so agents that load skills their own way keep working untouched. Native \texttt{@} integration stays small because it reuses the \texttt{@} context system every modern agent already has, and the host needs no protocol logic at all: the client computes the resident prompt block and the host splices in one string (\cref{app:protocol}).

\input{tex/fig_hero}

\minisection{The hub.} A file tree addresses skills well but cannot find them, host what has no repository, or show anyone what they have, so the protocol pairs with a free hub (\url{https://atskills.one}) whose value is deliberately confined to exactly those jobs. It \emph{finds}: search and ranking across the whole public corpus, which no amount of path grammar provides, with every result a copyable \texttt{@skills:} reference. It \emph{hosts}: a skill authored on the hub needs no GitHub repository at all, which matters because publishing today effectively requires maintaining one, and that keeps authorship technical even though the artifact is plain Markdown anyone could write. Private and team collections are hosted the same way, so work that will never be published gets the same addressing, sharing, and portability as work that is---and a personal collection travels with the user across projects, machines, and agents rather than living in a directory on one laptop. It \emph{manages}: a visual library of what a person or team owns, with usage visible per skill, where the file tree can only show a project one folder at a time. And it \emph{authors}: one screen, no repository, no git, for the non-developers whose procedural knowledge skills capture best. The division of labor is that GitHub hosts, the hub finds, and \texttt{@} uses. The git and GitHub precedent is the governing answer to why the hub can never be required: \texttt{gh:} paths and local folders resolve with zero hub involvement, forever, and GitHub-hosted skills keep their \texttt{gh:} identity even when the hub indexes them, since only skills authored on the hub carry its name. The hub earns its place on management value rather than lock-in, which is the same deal GitHub took.

\section{Discussion and Conclusion}
\label{sec:conclusion}

\minisection{The fix is subtraction, not addition.} The most useful thing to say about this protocol is how little of it is new. It adds no field to \texttt{SKILL.md}, no manifest, no lockfile, no database, and no service anyone is required to use. Nearly every piece already existed and is simply being used for delivery: the skill directory as the unit of content, the \texttt{@} gesture users already type to add files, \texttt{.gitignore} semantics for saying what fires on its own, git for transport and for revisions, vendoring for why a saved copy answers its own address, and the filesystem itself as the manager. What the protocol mostly does is take things away---the install lifecycle, the update command, version pinning, per-agent directories, and the whole packaging layer above the skill---and what remains is a path, a folder, and a file of one-line decisions. A user learns one character. An agent builder adds one file, or one dependency.

That matters because the competing convention won on exactly this axis. \texttt{AGENTS.md} became dominant not by being better written but by being trivial to deliver, and anything meant to succeed it has to be at least as simple to use, not merely better designed (\cref{app:discussion}).

\minisection{What it changes for the ecosystem.} For users, a skill can be tried once without being paid for on every message afterwards, so the decision to reach for one stops being a commitment. For authors, this is the difference between album sales and plays: today a skill outside a user's small installed set is never installed, so the long tail measures zero and its authors stop publishing, whereas usage counted per reference gives every published skill a real chance of being used. For teams, private work finally gets what public work has---one address, one place, reviewed like code and arriving with \texttt{git clone}---rather than an ecosystem built entirely for distributing to strangers. And for agent builders, a shared, agent-neutral store means a team stops maintaining the same skills once per agent, which is the cost that made working across agents impractical.

\minisection{Limits.} The protocol is additive, so installs, plugins, and vendor directories keep working, and nothing here asks a user to migrate. Our central quantity, the number of reliable auto-trigger slots, is bounded by argument and by the literature rather than measured by us, and the corpus figures come from one crawl and one working setup. On security, a referenced skill is remote text an agent will act on, which is the channel indirect prompt injection uses \citep{greshake2023indirect,liu2024injectionbench}; the answer is provenance and review rather than filtering, and the tiers form a trust ladder from a single throwaway session, to a git diff on save, to a one-line diff before anything joins the resident prompt. The risk is not new---an installed skill's body loads invisibly---but it is not removed either.

\minisection{Future work.} The measurements this argument invites are trigger reliability as a function of installed-skill count, and compliance as a function of injection position, both for skill-sized instructions; alongside them, the full classification of all 56{,}804 skills that \cref{app:landscape} begins, and reference-based usage counting in a public catalog, which is the direct test of whether authors publish more when plays rather than installs are the measure of use.

The principle generalizes past skills: \emph{resident context is a budget; spend it only on what must fire implicitly, and deliver everything else at the point of use, where attention is highest.}

\vspace{1mm}
\begin{center}
\emph{Install less, use more.}
\end{center}

\bibliography{main}

% ── Appendices: the full analysis, protocol, and evidence ──
\appendix
\crefalias{section}{appendix}
\section{Background: The Agent Skills Ecosystem}
\label{app:background}

\begin{quote}
\itshape A skill is a directory containing a \texttt{SKILL.md} file: organized folders of instructions, scripts, and resources that give agents additional capabilities.
\normalfont\hfill ---the official definition \citep{anthropic2025skills,anthropic2025skillseng}
\end{quote}

\minisection{From MCP to skills.} To see why skills spread so fast, and where their delivery model still falls short, start with what preceded them. The Model Context Protocol \citep{anthropic2024mcp} standardized how agents connect to tools, as a client--server JSON-RPC protocol, and won near-universal support within six months of its November 2024 release, adopted by OpenAI and Google alike. But MCP is heavyweight on both sides. The provider must build and maintain a server against a full specification covering transports, authentication, tools, resources, and prompts, and the measured quality of third-party servers is poor: in a study of 1{,}899 open-source MCP servers, 66\% exhibited code smells and 5.5\% shipped tool-poisoning vulnerabilities \citep{hasan2025mcpfirstglance}. Any open contribution ecosystem shares that hazard, and \cref{app:discussion} discusses the skills analogue. The user pays a standing cost per integration: a separately configured server process, tool schemas resident in context before any work begins, and every intermediate tool result passing through the model. GitHub's official server alone consumes tens of thousands of tokens across its $\sim$93 tool definitions \citep{willison2025skills}. By November 2025, MCP's own steward was steering integrations away from direct tool calls and toward code execution over the filesystem, reporting a drop from 150{,}000 to 2{,}000 tokens (98.7\%) on an equivalent task, with \texttt{SKILL.md} files as the durable, reusable layer \citep{anthropic2025codeexec,miessler2025mcpskills}. A skill, by contrast, is a folder of text the agent simply reads: no server, no protocol machinery, a fraction of the token cost \citep{willison2025skills}. Providers followed the simpler artifact \citep{kellner2026mcpskills}. Rather than maintain an MCP server, a provider ships a CLI plus a skill that teaches any agent to drive it. Cloudflare (\texttt{wrangler}), Expo (\texttt{eas}), Stripe, and Sentry all do this today \citep{cloudflare2026skills,expo2026skills,stripe2026skills,sentry2026skills}: one artifact, universal because virtually every coding agent has a shell.

\minisection{The \texttt{SKILL.md} format.} A skill is a directory containing a \texttt{SKILL.md} file: YAML frontmatter with a \texttt{name} and a one-line \texttt{description}, followed by Markdown instructions, plus optional scripts and reference files \citep{anthropic2025skills,anthropic2025skillseng}. The format is an open standard and deliberately minimal. A skill is readable by humans and by any model without tooling, and that simplicity is widely credited for its rapid uptake, since the same file works across harnesses with no protocol machinery at all \citep{willison2025skills,osmani2026skills}. Within a single loaded skill, the format already practices progressive disclosure: metadata stays resident, the body loads on relevance, and bundled files load on demand \citep{anthropic2025skillseng}. This paper extends that principle from \emph{within one installed skill} to \emph{delivery across the whole ecosystem}.

\minisection{What skills now carry.} The format was introduced for capabilities an agent lacked, but what people put in it has broadened, along two axes of ownership. \emph{Publicly}, a skill is either domain expertise offered to anyone, or a provider's integration path into every agent at once: rather than maintain a server per client, Cloudflare, Expo, Stripe, and Sentry each ship a CLI and a skill that teaches any agent to drive it \citep{cloudflare2026skills,expo2026skills,stripe2026skills,sentry2026skills}. \emph{Privately}, a team encodes its own way of working, which is knowledge that is valuable precisely because it is not public and will never reach a market. The two carry different value and demand different handling, and the second is the case existing tooling was never built for: a registry ranks by install count, and a team's deploy checklist has no install count. \Cref{app:failures} shows what fills the gap instead.

\begin{figure}[t]
\centering
\begin{tikzpicture}[font=\small, line join=round,
  chip/.style={draw=blue!40, fill=blue!8, rounded corners=2pt, inner sep=3.5pt, font=\scriptsize\ttfamily},
  dirbox/.style={draw=blue!40, fill=blue!6, rounded corners=2pt, inner xsep=6pt, inner ysep=5pt, align=left, font=\scriptsize\ttfamily, text width=2.3cm},
  note/.style={font=\scriptsize\itshape, text=black!45}]

  % ══ Left panel: agent configuration ══
  \draw[fill=black!3, draw=black!35, rounded corners=4pt] (0,0) rectangle (5.6,5.0);
  \node[anchor=west, font=\small\bfseries, text=black!70] at (0.3,4.6) {Agent configuration};

  % system prompt
  \draw[fill=white, draw=black!40, rounded corners=3pt] (0.35,3.2) rectangle (5.25,4.2);
  \node[font=\scriptsize, text=black!75] at (2.8,3.85) {Core system prompt};
  \node[note] at (2.8,3.48) {always in context};

  % installed skill descriptions
  \draw[fill=white, draw=black!40, rounded corners=3pt] (0.35,0.5) rectangle (5.25,2.85);
  \node[font=\scriptsize, text=black!75] at (2.8,2.55) {Installed skills --- \emph{descriptions}};
  \node[chip] at (1.1,1.95) {pdf};
  \node[chip] at (2.25,1.95) {docx};
  \node[chip] at (3.95,1.95) {nda-review};
  \node[chip] at (1.3,1.35) {bigquery};
  \node[chip] at (2.7,1.35) {xlsx};
  \node[chip] at (3.65,1.35) {\ldots};
  \node[note] at (2.8,0.82) {resident on every message};

  % ══ Right panel: agent file system ══
  \draw[fill=black!3, draw=black!35, rounded corners=4pt] (9.2,0) rectangle (15.2,5.0);
  \node[anchor=west, font=\small\bfseries, text=black!70] at (9.5,4.6) {Agent file system};

  \node[dirbox, anchor=north west] at (9.55,4.25)
    {skills/pdf/\\ \textcolor{black!55}{- SKILL.md}\\ \textcolor{black!55}{- reference.md}\\ \textcolor{black!55}{- extract.py}};
  \node[dirbox, anchor=north west] at (12.4,4.25)
    {skills/docx/\\ \textcolor{black!55}{- SKILL.md}\\ \textcolor{black!55}{- ooxml/}\\ \textcolor{white}{.}};
  \node[dirbox, anchor=north west] at (9.55,2.25)
    {skills/nda-review/\\ \textcolor{black!55}{- SKILL.md}};
  \node[dirbox, anchor=north west] at (12.4,2.25)
    {\ldots\\ \textcolor{white}{.}};
  \node[note] at (12.1,0.45) {bodies on disk, loaded only on trigger};

  % ══ Arrows in the gap ══
  \draw[-{Stealth}, black!35, dashed, thick] (9.2,3.85) -- (5.6,3.85);
  \node[font=\scriptsize, text=black!55, align=center] at (7.4,4.35) {install:\\[-1pt]copy dir, description $\to$ prompt};

  \draw[-{Stealth}, black!60, very thick] (5.6,1.5) -- (9.2,1.5);
  \node[font=\scriptsize, text=black!55, align=center] at (7.4,2.0) {auto-trigger:\\[-1pt]body loads on match\\[-1pt](probabilistic)};
\end{tikzpicture}
\caption{How installed skills work today (simplified from \citealp{anthropic2025skillseng}). Installation copies the skill directory onto the agent's file system and places its one-line description ($50$--$280$ tokens in our measured setup) permanently in the system prompt. The body---\texttt{SKILL.md} plus scripts and references---stays on disk and loads only when the model's probabilistic matching decides the description is relevant. Every installed skill pays the resident description cost on every message; whether it ever fires is left to the trigger lottery (\cref{app:failures}).}
\label{fig:install}
\end{figure}
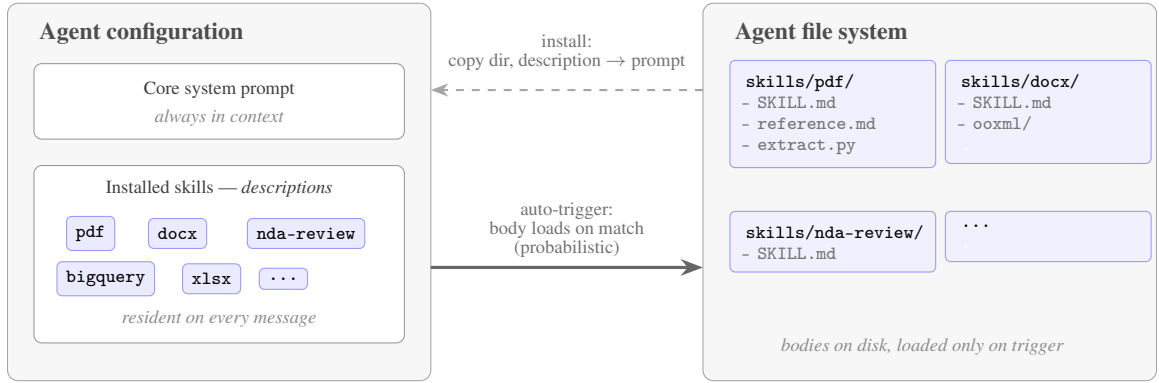

\minisection{Installation.} Agents take on skills by copying them into an agent-specific directory such as \texttt{.claude/skills/}, \texttt{.windsurf/skills/}, or \texttt{.adal/skills/} in AdaL, our own agent (\cref{app:protocol}). Across the 75 agents the community skills CLI supports there are 54 distinct project-level locations for the same format (\cref{app:landscape}). Users can install and uninstall, but each operation costs two to four CLI commands or app clicks, once per agent and once per machine. That overhead defeats what skills are for: knowledge that is simply \emph{there} when needed. At session start the agent puts each installed skill's name and description into the system prompt, and the model is expected to match requests against those descriptions and load the full body when one seems relevant. We call this \emph{auto-triggering}. The description is resident on every message; the body loads only on a trigger (\cref{fig:install}).

\minisection{No shared terminology, and management to match.} Above the skill sits a packaging layer, and there is no agreed definition of what it contains. In Claude Code, a plugin is a directory that bundles skills together with subagents, hooks, and MCP servers, and a marketplace is a catalog of plugins, so skills are \emph{children} of plugins: the desktop application lists installed plugins with a count of the skills inside each, and its composer reaches individual skills only through the plugin submenu \citep{claudecode2026pluginsref,claudecode2026plugins}. In the Codex application, plugins and skills are \emph{siblings}: a single settings pane titled ``manage plugins, skills, and MCPs'' presents plugins, apps, MCP servers, and skills as four parallel lists, and a plugin there is an installable bundle that may contain skills, service connectors, or both \citep{openai2026codexplugins}. In Cursor, skills are one of seven peer categories, alongside plugins, MCP servers, subagents, rules, commands, and hooks \citep{cursor2026customize}. The same word denotes a different object in each product, and the nesting is inverted between two of them. The only term that means the same thing everywhere is the skill itself.

Management inherits the confusion. Every agent renders its own inventory through its own interface, over its own hierarchy, in its own vocabulary, so a user running three agents learns three mental models and three management surfaces for what are, on disk, the same Markdown directories. Some surfaces are thin: an agent may show which bundles are installed without offering a view of the individual skills inside them, or a way to act on one. And because this machinery grew out of publishing, it is aimed at public distribution, which serves a provider shipping an integration and does nothing for a team whose skills will never go to a market. Teams do group their own skills as a collection grows, but that is an organizational convenience rather than a unit of delivery. The protocol in this paper keeps the one portable unit, the skill, as the thing users act on, and leaves public bundle distribution to the existing plugin and marketplace path (\cref{sec:hub}).

\minisection{Distribution.} The dominant registry, skills.sh \citep{vercel2025skills}, indexes public GitHub repositories and ranks skills by install count, and installation runs through a CLI (\texttt{npx skills add owner/repo}). A companion discovery convention lets any website self-host skills at \texttt{/.well-known/agent-skills/index.json} with content digests \citep{agentskills2025discovery}, though it is rarely used in practice, with Stripe the most prominent adopter. Publishing a skill therefore means, almost without exception, maintaining a GitHub repository. That keeps authorship technical even though the artifact is plain Markdown anyone could write (\cref{sec:hub}). Our July 2026 crawl resolves the indexed corpus to 56{,}804 \texttt{SKILL.md} directories across 1{,}133 public repositories. We use this directory count throughout as the size of the ecosystem. It is an upper bound on unique skills, because aggregator mirrors and per-agent duplicates are deduplicated only in the full-corpus analysis (\cref{app:landscape}).

\minisection{The competing convention: \texttt{AGENTS.md}.} In parallel, a much cruder mechanism has become the de facto standard for project-level agent instructions: a single \texttt{AGENTS.md} (or \texttt{CLAUDE.md}) file at the repository root, loaded in full on every message \citep{agentsmd2025}. It offers no modularity and no sharing beyond the repository, and it charges full token cost for every instruction on every message. Yet by widespread practitioner report it is the most widely adopted way to deliver agent knowledge. What fills it is the telling part. These files carry the project's and the team's \emph{workflows}: deploy procedures, review checklists, testing conventions, exactly the procedural knowledge skills were designed to package for a team. The role skills were built to play, a monolithic file plays instead. \Cref{app:failures} explains why, and \cref{app:discussion} returns to what its dominance implies.

\section{The Skills Landscape: What Skills Have Become}
\label{app:landscape}

Before analyzing how skills are \emph{delivered}, we examine what the ecosystem actually \emph{publishes}. Our July 2026 crawl resolves 56{,}804 \texttt{SKILL.md} directories across 1{,}133 public repositories; 56{,}245 skill bodies (99\%) from 1{,}132 repositories were fully fetched and analyzed. Four findings shape the rest of the paper. Two measurement regimes run throughout, and we mark which is which at every figure: fetch-based measurements (counts, body sizes, repository concentration) cover the full corpus, while classification figures (category shares, external-service requirements) come from a 100-skill uniform sample carrying roughly $\pm$10 percentage points at 95\% confidence. The structural findings below rest on the full-corpus measurements; the sampled classification corroborates them rather than carrying them.

\minisection{Finding 1: Skills are becoming a primary provider-to-agent integration channel.} The clearest structural signal in the corpus is who is publishing. Beyond individual authors and aggregators (repositories that re-host large collections of others' skills), \emph{first-party service providers} now maintain official skills repositories: AWS (\texttt{aws/agent-toolkit-for-aws}, 138 skills), Google (\texttt{google/skills}, 92), Elastic (\texttt{elastic/agent-skills}, 70), Grafana (49), LaunchDarkly (49), Sentry (80 across three repositories), Hugging Face (26), Expo (23), HashiCorp (17), Cloudflare (13, including a skill named simply \texttt{wrangler}), and Stripe, alongside Anthropic's own official plugin collections (379). In total we identify 21 major-provider organizations publishing 958 first-party skills, and a naming convention (\texttt{<org>/agent-skills} or \texttt{<org>/skills}) hardening into the de facto address for ``how agents drive our product.'' The content confirms these are integration artifacts rather than prose. Of the 569 skills in the fifteen provider \emph{product} repositories, 79\% mention the provider's own CLI or API tooling by name (\texttt{aws}, \texttt{gcloud}, \texttt{wrangler}, \texttt{eas}, \texttt{terraform}, \texttt{sentry-cli}, \ldots) and 60\% contain executable shell blocks.\footnote{Method: per-organization term lists matched as substrings over skill bodies; shell blocks detected by fenced \texttt{bash}/\texttt{sh} code or \texttt{npx}/\texttt{curl} usage. These detect mention rather than verified execution and carry some false positives; word-boundary matching and hand validation would tighten the estimate, and we report it as an upper bound.} The broader classified sample agrees: 42\% of skills require an account with an external service, and a further 26\% wrap a free tool. This is the corpus-level view of the provider migration described in \cref{sec:intro}. Instead of maintaining an MCP server per agent, a provider ships one CLI plus one skill and reaches every agent that has a shell \citep{kellner2026mcpskills,cloudflare2026skills,expo2026skills,stripe2026skills,sentry2026skills}. Two smaller specimens are equally telling. Stripe publishes the \emph{same} six skills duplicated under per-agent directories (\texttt{providers/claude/plugin/skills/}, and so on), which is the per-vendor packaging fragmentation a shared delivery protocol removes. And its flagship skill's one-line ``description'' runs to roughly 150 words of enumerated trigger conditions: the trigger-engineering pathology of \cref{app:failures}, practiced by a major provider.

\minisection{Finding 2: Skills are far too large to be resident, and just small enough to be fetched.} The median fetched skill body is 921 words and the 90th percentile is 2{,}207 (mean 1{,}159): on the order of one to four thousand tokens at the median, depending on tokenizer.\footnote{At 1.3 to 4 tokens per whitespace-delimited word, spanning typical prose ratios to a code-heavy upper bound, the 921-word median body is roughly 1.2k--3.7k tokens and the 90th percentile 2.9k--8.8k. We report the range rather than a point estimate because the exact figure is tokenizer-dependent; the argument here needs only the order of magnitude, which holds across the range.} Keeping even a modest working set of skill \emph{bodies} resident is out of the question. That is exactly why install-only delivery falls back to keeping only \emph{descriptions} resident and gambling on auto-triggering (\cref{app:failures}). The same numbers cut the other way, though: a few thousand tokens fetched once, at the point of use, is cheap. The corpus itself argues for on-demand delivery.

\minisection{Finding 3: The integration layer is fragmented per agent---54 directories for one format.} The \texttt{SKILL.md} \emph{format} is universal; where skills \emph{live} is anything but. The community skills CLI keeps an agent registry, the code that must know every agent's install location in order to work at all. It supports 75 coding agents using \textbf{54 distinct project-level skills directories}. Nineteen of the 75 have converged on \texttt{.agents/skills/}; the remaining 56 spread across 53 vendor-specific dotdirs such as \texttt{.claude/skills/}, \texttt{.windsurf/skills/}, \texttt{.devin/skills/}, and \texttt{.goose/skills/}, including, we must admit, our own \texttt{.adal/skills/} \citep{vercel2025skills}. User-level storage is worse, at 58 distinct home-directory locations across the same agents. The CLI copes the only way it can, installing once and symlinking into every detected agent's directory. The fragmentation is pure accident: every one of these directories holds the \emph{same} file format and differs only in whose namespace it sits in. A team running two agents maintains two skill installations; an author documents $N$ install paths; a repository accumulates one dotdir per agent its contributors run. \Cref{app:tiers} returns to this layer. The short version is that \atskill{} unifies it mostly by removing storage rather than by standardizing it.

\minisection{Finding 4: The long tail already lives in registries, not in installs.} Publication is heavily concentrated: the two largest aggregator registries alone account for 11{,}894 of the 56{,}245 fetched skills (21\%), and the top fifteen repositories for 51\%. The ecosystem has already solved \emph{hosting} and \emph{indexing}. Tens of thousands of skills sit one HTTP request away, while delivery stays gated on per-user installation. The distribution infrastructure exists; the delivery protocol is the missing piece.

\minisection{What a full-corpus classification would add.} The four findings above are what the corpus establishes. A classification pass over the entire 56{,}804-skill inventory, rather than the 100-skill sample, would sharpen five quantities without disturbing them: (i) the share and growth of provider-official skills, and how many wrap a CLI rather than a per-agent integration; (ii) the prevalence and token cost of trigger-engineered descriptions (descriptions padded with invocation phrases to win the auto-trigger lottery); (iii) exact token-cost distributions of descriptions and bodies under a production tokenizer; (iv) the category structure of what the world publishes, where our sample gives 30\% software engineering, 25\% creative production, 16\% writing, and 15\% design, with a long tail of business functions; and (v) duplication and staleness across aggregators. We flag these as refinements of magnitude rather than open questions of direction: each measures how strongly the ecosystem's binding need is a delivery protocol rather than more install tooling, and none of the paper's claims turn on where in the confidence interval they land.

\minisection{The landscape, distilled into a definition.} The official definition (\cref{app:background}) says what a skill \emph{is}. The corpus shows what skills are \emph{for}. We therefore extend it:

\begin{quote}
\itshape A skill is procedural knowledge for an AI agent, either the world's expertise or a team's own way of doing things, and it serves three roles: \emph{domain knowledge}, expertise distilled into instructions; \emph{provider integration}, by which a service teaches any agent to drive its product; and \emph{team workflows}, the customized procedures a team adapts to its own way of working. A model knows what the world knows; a skill adds what it doesn't.
\end{quote}

The rest of the paper is about giving each of these three roles the delivery it needs.

\section{Why Install-Only Delivery Fails}
\label{app:failures}

The failure is architectural: install-only delivery makes prompt residency plus probabilistic matching the \emph{only persistent} channel through which any of 56{,}804 published skills can reach an agent. The non-installing path now offered alongside it (\cref{sec:intro}) delivers a single named skill per invocation and caches nothing, so it relieves none of what follows: anything that must be there next session, anything at collection granularity, and anything whose cost is paid per message still goes through installation. Nothing in this section indicts auto-triggering itself; implicit activation is genuinely valuable, and the tier model retains it for exactly the instructions that need it (\cref{app:tiers}). Installed skills can also be invoked by name in the prompt, but that deterministic path exists only for the handful already installed and already paying residency; install-only delivery offers no explicit path to the rest of the corpus. This section measures the channel at three levels: it is physically scarce; authors and users rationally route around the scarcity; and the ecosystem inherits the damage. Six mechanisms, three levels, one cause.

\subsection{Level 1: The channel is physically scarce}
\label{sec:failure-channel}

Three independent mechanical properties of resident context, each documented in the literature, jointly cap what the channel can carry.

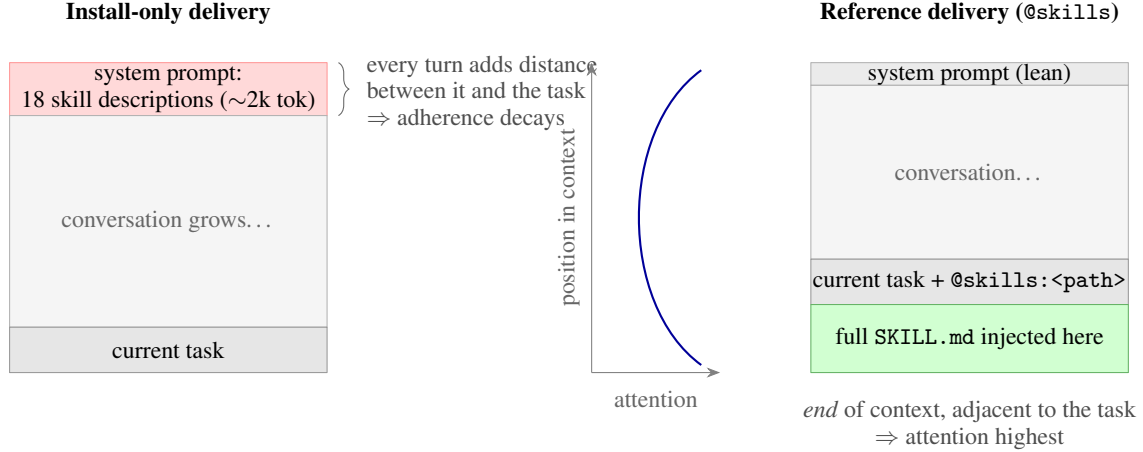
\begin{figure}[t]
\centering
\begin{tikzpicture}[font=\footnotesize, line join=round, lbl/.style={align=center}]

  % ---------- Left column: install-only ----------
  \node[lbl] at (0,4.75) {\textbf{Install-only delivery}};
  \draw[fill=red!15, draw=red!40] (-2.1,3.4) rectangle (2.1,4.1);
  \node[lbl] at (0,3.75) {system prompt:\\ 18 skill descriptions ($\sim$2k tok)};
  \draw[fill=black!4, draw=black!30] (-2.1,0.6) rectangle (2.1,3.4);
  \node[lbl, text=black!60] at (0,2.0) {conversation grows\ldots};
  \draw[fill=black!10, draw=black!40] (-2.1,0.0) rectangle (2.1,0.6);
  \node[lbl] at (0,0.3) {current task};
  \draw[decorate, decoration={brace, amplitude=4pt}, black!60] (2.25,4.1) -- (2.25,3.4);
  \node[anchor=west, align=left, text=black!70] at (2.5,3.7)
    {every turn adds distance\\ between it and the task\\ $\Rightarrow$ adherence decays};

  % ---------- Shared attention curve ----------
  \begin{scope}[xshift=5.6cm]
  \draw[-{Stealth[length=1.8mm]}, black!50] (0,0.0) -- (0,4.1);
  \node[rotate=90, text=black!60] at (-0.3,2.05) {position in context};
  \draw[-{Stealth[length=1.8mm]}, black!50] (0,0.0) -- (1.7,0.0);
  \node[text=black!60] at (0.85,-0.35) {attention};
  \draw[thick, blue!60!black] (1.45,4.0) .. controls (0.35,3.2) and (0.35,0.9) .. (1.45,0.1);
  \end{scope}

  % ---------- Right column: reference ----------
  \begin{scope}[xshift=10.6cm]
  \node[lbl] at (0,4.75) {\textbf{Reference delivery (\texttt{@skills})}};
  \draw[fill=black!8, draw=black!30] (-2.1,3.8) rectangle (2.1,4.1);
  \node[lbl] at (0,3.95) {system prompt (lean)};
  \draw[fill=black!4, draw=black!30] (-2.1,1.5) rectangle (2.1,3.8);
  \node[lbl, text=black!60] at (0,2.65) {conversation\ldots};
  \draw[fill=black!10, draw=black!40] (-2.1,0.9) rectangle (2.1,1.5);
  \node[lbl] at (0,1.2) {current task + \texttt{@skills:<path>}};
  \draw[fill=green!18, draw=green!50!black!60] (-2.1,0.0) rectangle (2.1,0.9);
  \node[lbl] at (0,0.45) {full \texttt{SKILL.md} injected here};
  \node[lbl, text=black!70, anchor=north] at (0,-0.25)
    {\emph{end} of context, adjacent to the task\\ $\Rightarrow$ attention highest};
  \end{scope}
\end{tikzpicture}
\caption{Delivery position is the mechanism. Left: an installed skill's description stays fixed at the top of the context, so the growing session pushes the task ever further from it, and adherence to it decays with the distance \citep{liu2024lost,qin2024sysbench}. Right: an explicit reference injects the full skill at the end of the context, adjacent to the task---the mechanically most reliable position an instruction can occupy. Center: attention as a function of position (schematic).}
\label{fig:position}
\end{figure}

\minisection{Distance decay: the request drifts away from the instruction.}
\label{sec:failure-position}
A resident description sits at a fixed place, the top of the context, and it stays there. What changes is the distance to the request it is supposed to match: every turn of conversation, every file read, and every tool result lands between the two, so an instruction that started next to the first request is separated from the tenth by everything that happened in between. Transformer language models attend most strongly to the beginning and the end of their context and under-weight what lies between \citep{liu2024lost,guo2024serial}. The bias is intrinsic and U-shaped rather than semantic, present regardless of how relevant the content is \citep{hsieh2024found}, and prompt-level fixes help only inconsistently \citep{guo2024serial}. Two lines of evidence measure the effect on instructions directly: adherence to the system message decays over conversation turns \citep{qin2024sysbench}, and models in long multi-turn settings lose an average of 39\% against single-turn performance on the same tasks \citep{laban2025lost}. The failure is positional, not semantic. The same instruction that goes unheeded from the top of a long session is followed when it arrives at the end of the context, next to the task (\cref{fig:position}). Explicit reference (\cref{app:tiers}) exploits exactly that.

\minisection{The standing tax: every install taxes every message.}
\label{sec:failure-tax}
Each installed skill's description stays resident permanently, paid on every message whether or not the skill is relevant. Measuring the installed skills on one real working setup, we find 50--280 tokens per skill; that session, with 18 installed skills, carries roughly 1{,}500--2{,}000 tokens of standing overhead on every request. The cost is not only financial. Input length alone degrades reasoning on an otherwise unchanged task, well below the model's nominal context limit \citep{levy2024same}; effective context is smaller than advertised context across model families \citep{hsieh2024ruler}; and prompt bloat is a recognized enough problem that an entire line of work exists to compress prompts back down \citep{jiang2023llmlingua}.

\minisection{Dilution: each additional description weakens the rest.}
\label{sec:failure-dilution}
More resident descriptions means more competition for the model's attention, and each individual skill fires \emph{less} reliably. Irrelevant context is not neutral---models are measurably distracted by it even when instructed to ignore it \citep{shi2023distracted}---and instruction-following degrades systematically with the number of simultaneous instructions: at 500 concurrent constraints even the best frontier model reaches only 68\% adherence, with degradation visible from far smaller counts \citep{jaroslawicz2025instructions}. The mechanism degrades exactly as adoption succeeds: the better the ecosystem does at getting skills installed, the worse each installed skill works. Bundling amplifies the problem, because the unit of installation is coarser than the unit of cost. Plugins are frequently loosely curated, and a single plugin can carry twenty or more skills; the dominant harness offers no per-skill control inside a plugin---install, enable, and disable all operate on the whole bundle \citep{claudecode2026plugins}. A user with a handful of plugins is quickly past a hundred resident descriptions, more entries than the agent's own built-in tool set, each one diluting the rest.

\minisection{The capacity: an attention budget of fewer than a hundred slots.}
\label{sec:failure-structural}
Jointly, the three mechanics cap what the channel can carry. We do not claim a precise constant; no controlled measurement of trigger reliability versus installed-skill count yet exists, and we propose one as future work (\cref{sec:conclusion}). But the evidence bounds the budget well under a hundred. Instruction-following degrades measurably as concurrent instructions accumulate, long before the hundreds \citep{jaroslawicz2025instructions}; that experiment measures concurrent constraints rather than trigger descriptions, so we take it as suggestive rather than direct. Practitioner guidance on resident context consistently counsels parsimony \citep{anthropic2025context}. We therefore claim only a conservative bound: \emph{fewer than one hundred reliable auto-trigger slots per agent}, and plausibly far fewer. The framing matters more than the digit: installation is an \emph{attention budget}, not a distribution channel---a physical property of resident context, not anyone's design mistake. Spending a slot is justified for an instruction that must fire unprompted; the pathology is that install-only delivery forces \emph{every} skill, however occasional its use, to bid for one. And wherever the true budget lies below one hundred, it is orders of magnitude smaller than 56{,}804.

\subsection{Level 2: Everyone rationally routes around the scarcity}
\label{sec:failure-behavior}

Given a scarce, probabilistic channel, every party adapts---and each adaptation makes the channel worse.

\minisection{Authors: trigger-engineering.}
\label{sec:failure-trigger}
An installed skill fires only if the model matches the user's message against its one-line description; matching competes with everything else in context, and misses are silent---the user never learns the skill existed. Authors respond rationally: examine popular published skills and the descriptions are dominated not by documentation but by \emph{trigger-engineering}:

\begin{quote}
\small\ttfamily TRIGGER --- read BEFORE opening the target file; don't skip because it ``looks like a one-liner''--- whenever: the prompt names [\ldots] in any form [\ldots] SKIP only when [\ldots]
\end{quote}

followed by paragraphs of trigger and skip conditions. This is a rational bid for a scarce slot, and it backfires collectively: among the examples we examined, TRIGGER-block descriptions cost roughly $20\times$ more tokens than quiet ones, feeding the tax and the dilution of \cref{sec:failure-channel} (full-corpus prevalence is part of the pending analysis, \cref{app:landscape}). Even major providers play this game: Stripe's flagship skill carries roughly 150 words of enumerated trigger conditions. Such blocks among widely-installed skills show authors pricing in, correctly, that the channel is unreliable.

\minisection{Users: installing is the wrong unit for trying something out.}
\label{sec:failure-equilibrium}
For a user, the cost of an install is certain and paid on every message (\cref{sec:failure-tax}), while the benefit is occasional and probabilistic. Installing is therefore worth it only for a skill that is used often, that clearly benefits from firing without being asked, and that is known to trigger reliably. Everything else falls outside those bars: skills being tried out, skills being compared against alternatives, skills needed once for an unusual task. For all of these, install-only delivery charges the full price of permanence. The user must place the skill, carry its resident description on every message thereafter, judge whether it is pulling its weight, and later find and remove it, in each agent and on each machine. That is heavy management for something the user wanted to \emph{try}, and the natural response is to not bother, which is how most of the corpus goes unexplored. Users do keep installing past that point, and the install itself is the easy part. What breaks down afterwards is memory: because the skill lives in a hidden per-agent directory and is meant to fire on its own, the user stops holding onto what is installed. When it then fails to fire (Level 1), nothing prompts them. They do not know a skill that would have helped was sitting right there, so they cannot invoke it by name either. Installation quietly transfers responsibility for remembering to a mechanism that only sometimes remembers. How many installs users actually settle at, and how trigger reliability falls as that number rises, are open empirical questions we propose measuring (\cref{sec:conclusion}).

\minisection{Teams: defection to \texttt{AGENTS.md}.}
Teams needing shared procedures skip the channel entirely: they paste workflows into a monolithic \texttt{AGENTS.md} (\cref{app:background})---always loaded, zero setup, no trigger lottery. The crudest delivery mechanism wins because it beats installation on the only axis users feel: simplicity. But the defection buys no reliability: \texttt{AGENTS.md} is resident context too, subject to exactly the distance decay of \cref{sec:failure-position}, hence the ubiquitous complaint that the agent ``ignores'' the team's instructed workflows, especially as the conversation grows long. The irony is sharp: a team's workflows are typically a manageable number---a few dozen at most---precisely the working set that skills were designed to carry, packaged one procedure per file. Teams thus end up, for lack of an alternative, on a mechanism that fails the same way. Adoption behavior has gone decisively against the install channel; \cref{app:discussion} returns to what that implies.

\subsection{Level 3: The ecosystem inherits the damage}
\label{sec:failure-ecosystem}

\minisection{The junk drawer: important skills get lost among clutter.}
\label{sec:failure-junkdrawer}
Because install is the \emph{only} lifecycle, everything lands in the same flat bin: a skill grabbed once for a one-off task sits forever next to the playbooks a team uses daily, and nothing distinguishes them. Users create skills expecting auto-trigger to remember for them, we conjecture, and so stop remembering themselves; then auto-trigger does not fire (Level 1), and the skill is simply forgotten. The consistently used, genuinely important procedures have no dedicated home and no management surface; they rot in a pile of once-used clutter. The resulting experience (install, forget, stop trusting, stop installing) is plausibly among the larger brakes on adoption of the entire ecosystem: a newcomer's first contact with skills is a directory of things that never fire, from which they conclude that skills do not work.

\minisection{The cap: distribution ends at each user's top handful.}
The structural consequence of all six mechanisms:

\begin{center}
\emph{56{,}804 indexed skills compete for fewer than 100 reliable auto-trigger slots per agent.}
\end{center}

The long tail has no delivery path; authors face zero measurable usage and stop publishing; users conclude the ecosystem is thin and stop looking. The install lifecycle---not the file format---is what strangles the ecosystem.

\section{The Three-Tier Delivery Model}
\label{app:tiers}

The failures of \cref{app:failures} share one root. Installation is the only tier, so every skill pays the same maximal cost, permanent prompt residency, for the same scarce benefit, a chance at auto-triggering, whether it is used once, weekly, or on every message. The fix is to let each skill sit in the \emph{cheapest tier that meets its activation need}. Splitting content, persistence, and triggering apart (\cref{sec:intro}) yields exactly three tiers (\cref{fig:tiers}). The three roles of \cref{app:landscape} map onto them directly: domain knowledge is the tier-1 long tail, team workflows are the tier-2 working set, and provider integrations reach agents through tier 1, plus tier 3 for the few that must fire unprompted.

\minisection{Tier 1: Reference.} \texttt{@skills:<path>} fetches a skill at the moment of use and injects its full body at the end of the context, next to the task, which is where attention is highest (\cref{sec:failure-position}). Triggering is explicit and therefore deterministic, the standing cost is zero, and the skill evaporates with the session. This tier serves the entire long tail: under reference delivery every published skill has a real chance of being used, because none of them has to win an install slot first.

\minisection{Tier 2: Saved.} A saved skill loads exactly like a reference, on demand and triggered explicitly, but it stays. It lives in a git-tracked directory in the repository, \texttt{.atskills/} in our implementation, where the team can find it, edit it, and version it alongside the code. Crucially, \emph{nothing} enters the prompt. The agent indexes \texttt{.atskills/} for \texttt{@} autocomplete and search, so saved skills surface the moment the user types, but that index lives in the tooling, outside the model's context. Zero resident tokens, zero trigger dilution, zero per-machine setup. This tier keeps the good half of installation, persistence and findability, and drops the bad half. It is per project rather than per machine, it arrives with \texttt{git clone} rather than through a teammate's install, it works offline, and resolution is instant because the files are already local. Tier 2 is the most underserved need in the current ecosystem, and it is where a team's \emph{working set} belongs: the deploy runbook, the review checklist, the migration procedure. It is also the better home for a fat \texttt{AGENTS.md}, the same instructions split into one procedure per skill, each loaded when invoked instead of taxing every message.

\minisection{Tier 3: Installed.} Installation stays the right mechanism for one class of skill: the ones that must fire \emph{without the user thinking of them}, such as formatter conventions, security guardrails, and API-reference checks. Even here, only the frontmatter, the name and one-line description, is resident; the body loads on trigger, exactly as the format prescribes (\cref{app:background}). Because auto-triggering is a bounded attention budget (\cref{sec:failure-structural}), tier 3 should stay small, and we recommend fewer than ten essentials, since every addition dilutes the rest. The protocol does not replace installation. It relieves it of the tens of thousands of skills that never belonged in the system prompt.

\minisection{Unified storage.} The three tiers map onto a unified storage story (\cref{fig:hero}), reusing existing conventions where they exist and introducing only what is missing:

\begin{itemize}
  \item \textbf{Tier 1} materializes nothing, or---for skills that ship scripts---writes into a session-scoped temporary directory, deleted after the session.
  \item \textbf{Tier 2} introduces one new location: \texttt{.atskills/} at the repository root, the persistent home of referenced skills a team decided to keep. As of July 2026, no agent claims this namespace. Saves are \emph{vendored} at the ID's own path---hub skills under their publisher's namespace (\texttt{.atskills/<owner>/<skill>/}), GitHub skills under \texttt{.atskills/gh/<owner>/<repo>/\ldots}---so skills from one source nest together, the tree stays readable at any size, and every saved copy answers its own address (\cref{sec:resolution}).
  \item \textbf{Tier 3} needs no new storage at all: install is \emph{one line} in \texttt{.atskills/.autotrigger}, a file with \texttt{.gitignore} semantics (\cref{sec:autotrigger}). A plain line auto-triggers a skill the project holds; an \texttt{@} line \emph{follows} a provider's skill without holding a copy; a trailing \texttt{/} takes a whole directory. The frontmatter stays the author's and activation becomes the consumer's, and one readable file shows the project's whole auto-trigger budget as one-line diffs any reviewer understands. Provenance needs no manifest either: a saved skill carries a two-line \texttt{.source} stamp, and a followed skill's provenance \emph{is} its line. Installation under the protocol is therefore project-scoped and git-tracked. The protocol defines no user-level install of its own and offers the hub instead (\cref{sec:hub}). Package management settled this scope question long ago: global installs lost to per-project dependencies, \texttt{npm -g} to \texttt{node\_modules} and system \texttt{pip} to virtual environments, because global state is invisible, forgotten, and unreproducible. A hidden home directory is where installs go to be forgotten, every agent renders it through its own interface, and the symlink bridge copies the same skill across 58 home-directory locations (\cref{app:landscape}). Existing vendor directories such as \texttt{.claude/skills/} and \texttt{.agents/skills/} stay readable for compatibility---the protocol feeds agents that load skills their own way; it does not replace them (\cref{sec:zero-integration}).
\end{itemize}

\minisection{User-level skills: the hub, not a dotdir.} One thing is deliberately absent: any \emph{new} user-level directory for the working set. We considered one and rejected it. A hidden dotdir outside the working tree is where skills go to be forgotten, since nobody revisits it and no editor or code review surfaces it, and it is awkward for developers and unreachable for everyone else. Personal and cross-project collections live in the hub instead (\cref{sec:hub}), managed in a dashboard rather than a dotdir. They lose nothing in immediacy, because the agent surfaces them directly: the \atskill{} autocomplete lists the user's cloud collection next to the project's saved skills, in any project and on any machine. The user gets the availability a home directory promises with the visibility it never delivers. The one thing personal skills give up is implicit activation, and we state that as a rule rather than a loss: \emph{auto-triggering is a project decision}. A personal skill that should fire automatically in some repository gets installed into that repository, one act and git-tracked (\texttt{:install}); everywhere else it stays one \texttt{@} away. Existing home-directory skills keep loading exactly as before. The protocol simply defines no user-level store of its own, and the hub is the alternative it offers, not a migration it forces.

\minisection{A lifecycle, not just a cost model.} The tier model is also a lifecycle, and that is what fixes the junk drawer (\cref{sec:failure-junkdrawer}). One-off skills never persist at all, because tier 1 evaporates with the session. Persistence happens only through \emph{deliberate promotion}: saving a referenced skill into \texttt{.atskills/} is an explicit act (\cref{app:protocol}), so the saved tree holds exactly what someone decided to keep. It is visible in the repository, reviewed like code, and maintained by the team, instead of an invisible per-machine directory collecting everything ever tried. Promotion is also the moment of \emph{adaptation}, since a saved copy can drift from its upstream source as the team fits it to their project. Through promotion and local edits, a generic public skill becomes the team's own practice.

The tiers complement each other rather than compete. A marketplace can still emit native install commands for the handful of essentials a user wants auto-triggered, while tiers 1 and 2 serve everything else ever published. Nothing asks the user to change: personal skills, plugin installs, and vendor directories keep loading as before, a project-saved copy deterministically shadows a same-name duplicate, and installing becomes one tier of three. The protocol does not replace the existing system. It \emph{completes} it.

The tier model is also where this design meets emerging industry guidance on context engineering: treat context as a finite resource with diminishing returns, and prefer just-in-time retrieval of lightweight identifiers over pre-loading content \citep{anthropic2025context}. The three tiers put that guidance into practice for skills, and go one step further. Tier 2's search index sits \emph{outside the context altogether}, in the tooling, so even the lightweight identifier costs nothing until the user types \texttt{@}. Tiers 1 and 2 \emph{are} just-in-time retrieval, with the piece the guidance lacks: an ecosystem-wide protocol and catalog behind the fetch (\cref{app:protocol}).

\minisection{What the user gains.} The benefits compound into a different experience of skills:

\begin{itemize}
  \item \textbf{Granularity.} Save or install \emph{one} skill instead of a twenty-skill plugin, so the unit of control finally matches the unit of cost (\cref{sec:failure-dilution}).
  \item \textbf{Transparency.} Every skill has exactly one knowable place: the project directory, a git diff away; the hub, a page with an owner and a history; or the session transcript. The project's whole auto-trigger budget is one readable file.
  \item \textbf{Clean prompt, free management.} The system prompt carries only the few essentials someone deliberately chose, and everything else costs nothing until the moment of use. Prompt hygiene stops being a chore because there is nothing to clean.
  \item \textbf{Robust manual triggering.} An explicit reference loads every time, at the best position for compliance, with no lottery and no silent misses.
  \item \textbf{Sharing.} Git carries the project set to every teammate on clone, the hub carries personal and team sets to every machine, and publishing needs no repository.
  \item \textbf{Unification.} One syntax, one project directory, one trigger file, one hub, and the same experience in every agent at either integration level.
\end{itemize}

\section{The \texorpdfstring{\atskill{}}{@skills} Protocol: Full Specification}
\label{app:protocol}

The protocol is deliberately small enough to state completely. It consists of four file-level mechanisms---a reference command (\texttt{@skills:<path>}), a project folder (\texttt{.atskills/}), a trigger file (\texttt{.autotrigger}), and a provenance stamp (\texttt{.source}, the two lines recording a saved copy's origin and the revision it was taken at)---plus a management surface (\texttt{/skills}) that every conforming client should ship. Throughout, the unit of content is the unmodified \texttt{SKILL.md} directory of \cref{app:background}: the protocol adds \emph{zero} fields to the format. It is developed in the open, with the specification, the agent instruction file (\texttt{SKILLS.md}), a reference CLI, and a runnable demo maintained at \url{https://github.com/SylphAI-Inc/atskills} \citep{agentatskills}. AdaL implements the protocol end to end---native references, autocomplete, the global validating cache, saves, the auto-trigger file, and the \texttt{/skills} checkbox tree---in both its CLI and browser application \citep{adal2026}. \Cref{app:algorithms} states the four normative algorithms; this section specifies the artifacts and their meaning.

A design invariant governs everything below: \textbf{the protocol is purely a filesystem}---folders, files, lines, nothing else. There is no manifest, no lockfile, no database, and no state that is not visible in an editor and reviewable in a pull request. A second invariant bounds the state space: there are exactly \textbf{two states}. \emph{Theirs}: a skill followed by one line in \texttt{.autotrigger}, never held locally---the provider keeps it true because it documents the provider's service. \emph{Yours}: a folder in \texttt{.atskills/}, either written in place or saved-to-adapt, detached from upstream at the moment of saving. The muddy third state of every install system---installed-but-stale, enabled-but-unknown-origin, update-pending---is designed out, not managed.

\subsection{Reference syntax and identity}
\label{sec:syntax}

A skill is referenced inline, in the user's message or by the agent itself. The \emph{path is the identity}:

\begin{lstlisting}
@skills:hub:sylphai/glowmotion              one skill, from the hub
@skills:gh:acme/skills/deploy           one skill, straight from GitHub
@skills:gh:stripe/agent-toolkit         a directory -> a menu, one line per skill
@skills:<path>:save                     copy into the project -- to adapt it
@skills:<path>:install                  append its line to .autotrigger
@skills:<path>:save:install             both -- your copy, firing on its own
\end{lstlisting}

Identity rules: hub paths (\texttt{hub:owner/name}) are lowercase, and resolvers fold case so it can never split an address; \texttt{gh:owner/repo/sub} keeps GitHub's casing beyond the marker, because GitHub paths are case-sensitive. Pasted GitHub URLs (\texttt{github.com/\ldots/tree/<branch>/\ldots}) normalize to the \texttt{gh:} form. On disk, \texttt{gh:} is spelled \texttt{gh/}, since folder names cannot hold colons. One skill has one ID, forever: GitHub-hosted content is always \texttt{gh:}, even when the hub indexes or serves it (\cref{sec:hub}).

The grammar is greedy: the path runs until the end of the token or the trailing suffixes. The two suffixes are \emph{orthogonal and combinable}---\texttt{:save} means own a copy, \texttt{:install} means one line in \texttt{.autotrigger}---and neither implies the other. Several references in one message all load, each at its own point of use, which is what makes multi-skill workflows deterministic: install-only delivery has no equivalent, since co-firing $N$ skills means $N$ descriptions winning a probabilistic match at once (\cref{sec:failure-dilution}).

\minisection{A directory is a menu.} A path with no \texttt{SKILL.md} is not a failure---it is a directory, and it loads as a menu: one line per skill under it, \texttt{path: description}, every line itself a valid reference. Browsing and using are the same gesture, and a bundle is only ever a menu: skills are taken one path at a time, and all-or-nothing delivery does not exist in the protocol. The menu is produced by the \emph{leaf rule}: a folder holding \texttt{SKILL.md} is a skill and the walk stops there, so nested repositories can never yield a skill inside a skill, and repository cruft (\texttt{README.md}, \texttt{LICENSE}, \texttt{docs/}) is never mistaken for content. Only a \texttt{SKILL.md} at the referenced path itself makes the reference a single skill.

\minisection{The autocomplete rule.} \texttt{@skills:} completes only what the project already knows: its local skills and the cloud IDs in \texttt{.autotrigger}. For the world, the user types or pastes a path---GitHub URLs work---and discovery is the hub's job, not the input box's. The learning curve is one character.

\subsection{Resolution: by path, through a validating cache}
\label{sec:resolution}

\begin{figure}[t]
\centering
\begin{tikzpicture}[font=\footnotesize, node distance=0.7cm and 0.9cm,
  stage/.style={draw=black!50, fill=black!4, rounded corners=2pt, align=center, inner sep=5pt, minimum height=0.9cm},
  good/.style={draw=green!50!black!60, fill=green!15, rounded corners=2pt, align=center, inner sep=5pt, minimum height=0.9cm},
  warn/.style={draw=orange!70!black!60, fill=orange!15, rounded corners=2pt, align=center, inner sep=5pt, minimum height=0.9cm},
  bad/.style={draw=red!60!black!60, fill=red!10, rounded corners=2pt, align=center, inner sep=5pt, minimum height=0.9cm},
  go/.style={-{Stealth[length=1.8mm]}, black!70}]

  \node[stage] (ref) {\texttt{@skills:<path>}};
  \node[stage, right=of ref] (local) {folder at\\ \texttt{.atskills/<path>}\,?};
  \node[stage, right=1.6cm of local] (probe) {the cloud:\\ changed? \emph{(one probe)}};

  \node[good, below=of local] (yours) {yours --- read it,\\ done};
  \node[good, below=0.7cm of probe, xshift=-2.2cm] (cache) {unchanged:\\ cache, instant};
  \node[good, below=0.7cm of probe, xshift=0.35cm] (fresh) {changed:\\ fetch fresh};
  \node[warn, below=0.7cm of probe, xshift=2.9cm] (stale) {offline, cached:\\ serve, mark stale};
  \node[bad, right=0.5cm of stale] (fail) {nothing cached:\\ fail, say why};

  \draw[go] (ref) -- (local);
  \draw[go] (local) -- node[left, text=black!60] {yes} (yours);
  \draw[go] (local) -- node[above, text=black!60] {no} (probe);
  \draw[go] (probe.south) -- (cache.north);
  \draw[go] (probe.south) -- (fresh.north);
  \draw[go] (probe.south) -- (stale.north);
  \draw[go] (probe.east) -| (fail.north);
\end{tikzpicture}
\caption{Resolution is by path, through a validating cache (\cref{alg:resolve}). A folder at \texttt{.atskills/<path>} is the project's own and always answers---a saved copy sits at its ID's own path, so it answers its own address with no lookup machinery. Otherwise the path means the cloud: one revision probe decides between serving the shared global cache instantly, fetching fresh, or (offline) serving the cache marked stale; every green outcome injects at the point of use. Oversized references---a whole marketplace repo as one path---are refused before any download, with the loadable sub-collections named.}
\label{fig:resolution}
\end{figure}
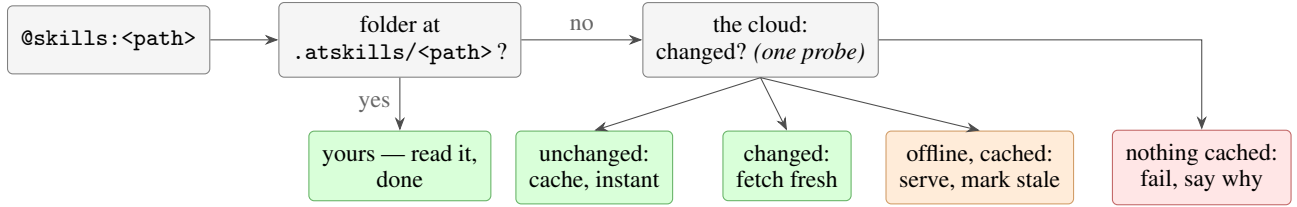

Resolution has one local rule and one cloud rule (\cref{fig:resolution}, \cref{alg:resolve}):

\begin{enumerate}
  \item \textbf{Local first, by path.} A folder at \texttt{.atskills/<path>} (\texttt{gh:} spelled \texttt{gh/}) is the project's own and always answers; read it, use it, stop. No folder there means the path means the cloud. Nothing else is consulted---in particular, \texttt{.source} (\cref{sec:save}) is \emph{never} read to resolve anything. Because a saved copy sits at its ID's own path, it answers its own address by construction: this is vendoring, with deep precedent---Go's \texttt{vendor/github.com/acme/\ldots} mirrors the import path and resolves first, node's \texttt{node\_modules/@scope/pkg} keeps the scope. Move or rename the folder and it simply stops answering the old address.
  \item \textbf{Else the cloud, through the global validating cache.} Cloud content materializes under one machine-wide, agent-neutral cache root (\texttt{\$XDG\_CACHE\_HOME/atskills/<disk path>}, defaulting to \texttt{\textasciitilde/.cache/atskills/<disk path>}), shared by every conforming client on the machine. This root \textbf{must not} live inside any \texttt{.atskills/} directory: a project whose root is the home directory would otherwise enumerate the machine's cache as its own skills, and an auto-trigger line written against a cached copy is git-tracked but machine-local, resolving to nothing on a teammate's checkout. The cache has browser semantics: each use asks the source ``did this change?''---a single revision probe, never a re-download---and \emph{unchanged} serves the cache instantly, \emph{changed} fetches fresh, \emph{unreachable} serves the cached copy with a stale warning, and unreachable-with-nothing-cached fails and says exactly why. Entries are always safe to delete; the path re-resolves.
\end{enumerate}

The reference transport for GitHub content is git itself: one shallow, blob-filtered, sparse clone of the referenced sub-path, and one \texttt{ls-remote} for the change probe. This choice is deliberate. Git negotiates the transfer in one round trip, needs no API quota (REST-based transports die at unauthenticated rate limits in practice), works against private repositories through the user's existing credentials, and gives revisions for free: the probe is a commit hash, and a pinned-revision fetch (needed by save-again verification, \cref{sec:save}) is a plain fetch-by-sha. A machine without git gets one clear error naming the missing tool---not a slower hand-rolled fallback.

\subsection{\texttt{.atskills/} and \texttt{.source}: save = adapt + detach}
\label{sec:save}

Everything under \texttt{.atskills/} is the project's: either written in place (any name, no stamp), or saved because someone intends to \emph{adapt} it. Saving is a download, not a subscription---the copy detaches from its provider the moment it lands. A provider skill nobody will edit does not belong here at all; it is followed with one line (\cref{sec:autotrigger}) and never held.

\begin{lstlisting}
.atskills/
|-- .autotrigger                 what fires on its own
|-- my-tdd/SKILL.md              yours -- any name, no .source
|-- hub/sylphai/glowmotion/          saved to adapt -- the path is the ID
|   |-- SKILL.md
|   `-- .source
`-- gh/stripe/agent-toolkit/     saved directory -- one .source covers it all
    |-- .source
    |-- payments/SKILL.md
    `-- terminal/SKILL.md
\end{lstlisting}

\texttt{.source} is two lines, written once at save time and never touched again:

\begin{lstlisting}
gh:stripe/agent-toolkit
2026-08-01 rev:abc123...
\end{lstlisting}

Line 1 is the origin ID; line 2 is the date and the upstream revision taken. It is pure provenance---a birth certificate, not a leash: the resolver never reads it, nothing syncs against it, and deleting it detaches fully. A subtree save writes \emph{one} stamp at the top of what was saved; the closest \texttt{.source} at or above a skill is its origin. Absence of a stamp is itself information: no \texttt{.source} means the project wrote it, and the cloud is never consulted about it.

\minisection{No update lifecycle---save-again instead.} A saved skill is detached on purpose, so there is nothing to ``update'': edit it, commit it, git carries the history. Curiosity about upstream is a question, not a lifecycle: the agent reads \texttt{.source}, fetches, and diffs, with line 2 separating ``you changed it'' from ``they changed it.'' Saving the same path again is the only refresh, and it is conflict-safe by construction (\cref{alg:save}): an \emph{unedited} copy is replaced and line 2 rewritten, where unedited is verified by re-fetching upstream \emph{at the recorded revision} (immutable, fetchable by hash) and comparing bytes---no content digests, no staging directories, no stored state beyond the two lines. An \emph{edited} (or unverifiable) copy is a conflict, and a conflict touches nothing and lists the ways out: keep yours (do nothing); refetch (delete the folder and save again---git keeps the history); or merge (ask the agent, with line 2 as the base). There is deliberately no three-way text merge in the protocol: reconciling two versions of prose instructions is a semantic judgment, and the judge is already in the terminal. Remove is deleting the folder. No other machinery exists.

\minisection{No pinning, ever.} The protocol defines no versions, releases, or lockfiles. A skill documents a living service; freezing the text while the API moves is false safety---old instructions for a world that did not freeze. The two honest relationships are \emph{follow} (the provider keeps it true) and \emph{own} (you hold the text and own keeping it true); real control over what the agent reads is a local copy, and that is exactly what save is for.

\subsection{\texttt{.autotrigger}: install is a line}
\label{sec:autotrigger}

One file governs everything that fires on its own, and it works like \texttt{.gitignore}: one entry per line, \texttt{\#} for comments, duplicates load once, order preserved.

\begin{lstlisting}
# .atskills/.autotrigger
sec-checklist                       yours -- lives in .atskills/, reviewed
team-flows/                         yours -- every skill under the directory
@hub:sylphai/glowmotion                 hub -- follows the author's latest
@gh:stripe/agent-toolkit/payments   github -- follows Stripe's latest
@gh:stripe/agent-toolkit/           github directory -- all of it
\end{lstlisting}

The line tells you everything: plain = your file, \texttt{@} = the cloud, \texttt{@gh:} = GitHub, trailing \texttt{/} = the whole directory. A plain line is a gitignore \emph{pattern} over the local skill tree---globs and \texttt{!} negation included, with all plain lines forming one ruleset so negation composes exactly as it does in git. \textbf{Install = a line in this file.} That is the whole equation: installing a skill is adding its line, uninstalling is removing it, and the suffixes, the \texttt{/skills} checkboxes, and hand edits in any editor all write the same lines.

What ``fires on its own'' means, precisely (\cref{alg:residency}): at session start each resident skill contributes its \emph{frontmatter only} (name + description, $\sim$50--100 tokens) to the always-available index; the body loads only on trigger---the format's own progressive disclosure. Cloud lines refresh once per session through the validating cache, so an unchanged upstream costs one revision probe; offline serves the last cached copy, marked stale. Every line resolves local-first, so a saved copy answers its own \texttt{@} line and what fires is the team's adaptation. Per-line failures are isolated: a line that loads nothing is reported once and the session goes on. A directory line's cost is its children's frontmatter, and the management surface always shows the expanded count and cost, never one opaque line; a directory line's drift includes entirely new skills arriving under it, and the count change is shown at session start.

\minisection{The trust trade, stated honestly.} An \texttt{@} line is reviewed once, as a one-line diff, and then the provider's updates flow in unreviewed---that is what following \emph{means}, and the management surface marks it. Scripts are confirmed \emph{by change, not by location}: the first run of a cloud skill's script shows the command and asks; it asks again only when the skill's revision changed since the last confirmed run (the cache already knows the revision). A followed provider skill therefore never nags, and saving is never needed just to silence confirmations. Follow what's theirs; save what you'll make yours.

\subsection{\texttt{/skills}: the management surface}
\label{sec:skills-surface}

Part of the protocol, not the product: non-technical users are first-class, so every conforming client should ship the management surface, and the honest claim is ``no state beyond the files,'' not ``no management surface.'' \texttt{/skills} is a view and editor over \texttt{.atskills/} and \texttt{.autotrigger}---nothing else:

\begin{lstlisting}
/skills
  [x] team-flows/                    check a whole directory
      [#] deploy
      [#] review-checklist
  [ ] my-tdd
  [x] @gh:stripe/agent-toolkit/payments   cloud - auto-updates
                                    budget: 4 skills - ~260 tokens
\end{lstlisting}

Check a box and a line is written; uncheck and it is removed. The checkbox states mirror the file faithfully: checked directly (\texttt{[x]}, its own line), covered by a directory line or pattern (\texttt{[\#]}), partially covered (\texttt{[\~{}]}). Unchecking one skill under a covering directory line performs a \emph{split}: the directory line is replaced by explicit lines for the siblings that stay on, so the file always reads true (\cref{alg:toggle}). Every operation is equally a typed command (\texttt{/skills save | install | uninstall | remove | toggle <path>}), and the surface shows each skill's origin (\emph{yours} / \emph{saved from\ldots} / \emph{cloud}) and the expanded cost of every line. The key affordance is \textbf{view prompt}: the exact text injected into the model, word for word, with its token count. You read what the model reads; no agent offers this today.

\subsection{Adoption Level 0: one file or one CLI, zero integration}
\label{sec:zero-integration}

The protocol asks nothing of the agent or its vendor---no SDK, no registry client, no plugin. \texttt{SKILLS.md}, one instruction file shipped in the protocol repository and droppable into any repository, turns any agent that can read files, run shell commands, and fetch URLs into a full client: it states the resolution rule, the cache discipline, the \texttt{.autotrigger} semantics, the save procedure, and the safety rules, in about a page. The agent \emph{is} the integration.

The same repository ships the second zero-integration path: a small dependency-free reference CLI. An agent that shells out to it inherits the cache, the resolution order, and the save rules without implementing any of them:

\begin{lstlisting}
atskills get <path>       resolve local -> cache -> web; print the files
atskills save <path>      copy to .atskills/<path>/ + .source
atskills triggers         parse .autotrigger; print what loads and its cost
atskills prompt           the exact resident block, word for word
atskills skills           the /skills management surface
\end{lstlisting}

Either way, a resolved skill is just files on disk. Agents that load skills their own way---\texttt{--skill} flags, native folders---keep working untouched: the protocol feeds them; it does not replace them.

\subsection{Adoption Level 1: native \texorpdfstring{\texttt{@}}{@} integration}
\label{sec:ladder}

Level 1 lives in the agent, and it is deliberately small because it reuses the \texttt{@} context system every modern agent already has: references are detected exactly where \texttt{@file} mentions already are, resolve by \cref{alg:resolve}, and stream content into the context at the point of use; autocomplete surfaces the project's local skills and its followed cloud IDs in the same dropdown users already use for files.

One architectural property makes Level 1 cheap and is worth stating as part of the protocol: \textbf{the host needs no protocol logic}. In AdaL's implementation, the frontend client owns everything---parsing \texttt{.autotrigger}, resolution, the validating cache, saves, and the construction of the resident prompt block---and the model-serving backend receives that block as \emph{one string} it splices into the system prompt verbatim. The protocol implementation therefore lives in exactly one codebase per client, the same one that renders the management surface, and the resolution the user sees in the dialog can never disagree with what the model receives. Any agent architecture with a ``prompt assembly'' step can adopt the protocol this way, whatever language its serving layer is written in.

\subsection{The hub: finding and authoring as a free service}
\label{sec:hub}

A protocol alone does not solve finding, so it pairs with a free hub (\url{https://atskills.one}) \citep{atskillshub2026}. The hub is a service, never a requirement---\texttt{gh:} and local paths resolve with zero hub involvement, and hub paths are plain HTTP GETs anyone can mirror. The governing precedent is git and GitHub: git is fully open, nobody is forced onto GitHub, and GitHub became indispensable by being the best place rather than the required place. Forcing the hub would kill both sides---a protocol that funnels users to one server is not a protocol, and an unadopted protocol brings the hub no crowd---so \texttt{gh:} stays first-class forever, and the hub competes only on the jobs GitHub is genuinely bad at: search over the 56k-skill corpus, one-screen authoring for the non-developers whose procedural knowledge skills capture best, private and team collections, and the user's own library.

Identity honesty is a hard rule: for GitHub-hosted skills the hub is a mirror plus search with enriched metadata---categorization, better descriptions, rankings---and nothing more; identity stays \texttt{gh:} even when the hub indexes, serves, or features the skill. Only skills authored on the hub carry the hub's own identity. The hub never renames other people's work, so the aliasing problem cannot exist. For curation the hub uses \emph{playlists}: named lists of \texttt{gh:} IDs, searchable and one \texttt{@} away, behaving exactly like directories everywhere else in the protocol---\texttt{@}-able as a menu, savable as a set, installable as one line---the way a music service links tracks it does not own.

\minisection{Bundles: granular by design.} The protocol deliberately defines no bundle-level delivery: the unit of delivery is a single skill because the unit of cost is a single skill (\cref{sec:failure-dilution}), and a directory or playlist is always a menu, never an all-or-nothing install. Plugin and marketplace installs are not removed; they keep working unchanged alongside the protocol for users who want bundles. Scope is bounded the same way: the protocol distributes \emph{documentation}. Plugin machinery that installs capability---hooks, MCP servers, slash commands---is a different problem and stays out; what dissolves here is the documentation half of ``plugins,'' which is most of it.

\section{The Protocol as Algorithms}
\label{app:algorithms}

The protocol is small enough to state completely. This appendix gives the four normative algorithms---resolution, save, residency, and the checkbox toggle---exactly as the reference implementations execute them \citep{agentatskills,adal2026}. Two properties are worth reading them for. First, \emph{statelessness}: every algorithm is a function of the filesystem (the project tree, the trigger file, the stamps, the cache) and the upstream; there is no database, no manifest, and no state that a user cannot inspect in an editor. Second, \emph{honest failure}: every refusal names its reason and, where one exists, the reference that would have worked.

Notation: $K = 128$ is the collection cap; $\mathrm{disk}(id)$ spells \texttt{gh:} as \texttt{gh/}; $\mathrm{walk}(d)$ enumerates skill directories under $d$ by the \emph{leaf rule} (a folder holding \texttt{SKILL.md} is a skill and the walk stops there, so a nested \texttt{SKILL.md} belongs to the bundle above it); $\bot$ denotes ``unreachable.''

\subsection{Resolution}

\begin{algorithm}[h]
\caption{\textsc{Resolve}$(p)$ --- local first, by path; then the validating cache}
\label{alg:resolve}
\begin{algorithmic}[1]
\State $id \gets \textsc{Normalize}(p)$ \Comment{fold hub case; \texttt{gh:} keeps casing; GitHub URLs $\to$ \texttt{gh:}}
\State $d \gets \texttt{.atskills/} \Vert \mathrm{disk}(id)$
\If{$d$ is a directory with content} \Comment{\textbf{local first} --- \texttt{.source} is never consulted}
  \State \Return \textsc{Describe}$(id, d)$
\EndIf
\State \Return \textsc{ReadThroughCache}$(id)$
\Statex
\Function{Describe}{$id, d$} \Comment{a skill, or a directory-as-menu}
  \If{$d/\texttt{SKILL.md}$ exists} \Return skill: body $+$ bundled files \EndIf
  \State \textbf{require} $|\mathrm{walk}(d)| \le K$ \Comment{else refuse, listing the largest \emph{loadable} sub-collections}
  \State $M \gets \{\, (id/rel,\ \mathrm{frontmatter}) : rel \in \mathrm{walk}(d) \,\}$
  \If{$M = \emptyset$} \textbf{fail} ``nothing at $id$'' \EndIf
  \State \Return menu $M$ \Comment{one line per skill; every line a valid path}
\EndFunction
\Statex
\Function{ReadThroughCache}{$id$} \Comment{browser semantics over git}
  \State $c \gets \mathrm{cache} \Vert \mathrm{disk}(id)$;\quad $m \gets \mathrm{meta}(id)$
  \State $r \gets \textsc{HeadRevision}(id)$ \Comment{one \texttt{git ls-remote} --- a probe, never a download}
  \If{$c$ cached \textbf{and} $r \ne \bot$ \textbf{and} $r = m.\mathrm{rev}$}
    \State \Return \textsc{Describe}$(id, c)$ \Comment{\textbf{unchanged}: served instantly}
  \ElsIf{$c$ cached \textbf{and} $r = \bot$}
    \State \Return \textsc{Describe}$(id, c)$ with stale warning \Comment{\textbf{offline}: cache, marked}
  \EndIf
  \State \textbf{try} \textsc{Fetch}$(id, c, r)$;\ $\mathrm{meta}(id) \gets (r, \mathrm{now})$;\ \Return \textsc{Describe}$(id, c)$
  \State \textbf{on failure:} serve the cached copy with a stale warning if one exists; else \textbf{fail} with the reason
\EndFunction
\end{algorithmic}
\end{algorithm}

\textsc{Fetch} for \texttt{gh:} content is git in three phases, and the order is the security and scale property. \emph{Phase 1, trees only}: a shallow, blob-filtered, no-checkout clone (\texttt{--depth 1 --filter=blob:none --no-checkout}, \texttt{--sparse} when a sub-path is given) transfers the directory structure but not one file body. \emph{Phase 2, count before paying}: \texttt{git ls-tree -r -l} lists every path with its blob size from tree metadata alone; the leaf-rule count over the referenced sub-path enforces $K$ \emph{before} any content downloads, and a refusal reports the reference's true weight (``6{,}296 skills, 1.7\,MB'') plus the largest sub-collections that fit---in a measured crawl, the 15 repositories over 500 skills hold 51\% of the public corpus, so the cap touches almost no repositories while defusing half the risk. \emph{Phase 3, materialize}: sparse-checkout of the sub-path, then the leaf-rule copy---\texttt{SKILL.md} at the top means the whole tree is that one skill's bundle; otherwise each skill folder lands at its true relative depth, and repository cruft (\texttt{README}, \texttt{LICENSE}, \texttt{.github/}), \texttt{.git} itself, and symlinks never land. The cap counts \emph{skills, never files}, so a single skill with a large bundle is always allowed.

\subsection{Save = adapt + detach}

\begin{algorithm}[h]
\caption{\textsc{Save}$(id)$ --- conflict-safe by construction, no stored state beyond two lines}
\label{alg:save}
\begin{algorithmic}[1]
\State $dest \gets \texttt{.atskills/} \Vert \mathrm{disk}(id)$
\If{$dest$ holds skills} \Comment{save-again: the only ``update'' in the protocol}
  \State $s \gets$ nearest \texttt{.source} at or above $dest$
  \If{$s = \bot$} \textbf{refuse}: it is the project's own work; nothing is touched \EndIf
  \If{$\lnot$\textsc{Unedited}$(id, dest, s.\mathrm{rev})$}
    \State \textbf{conflict}: touch nothing; offer \emph{keep} $\cdot$ \emph{refetch} $\cdot$ \emph{agent merge with base $s.\mathrm{rev}$}
  \EndIf
\EndIf
\State $r \gets \textsc{HeadRevision}(id)$
\State \textsc{Fetch}$(id, staging, r)$; \textbf{atomically} swap $staging \to dest$ \Comment{a failed fetch can never destroy a copy}
\State write \texttt{.source}: line 1 $= id$;\ line 2 $= \mathrm{date} + r$ \Comment{written once; never resolved against}
\Statex
\Function{Unedited}{$id, dest, rev$}
  \If{$rev = \bot$} \Return false \Comment{unverifiable counts as edited --- refusing is the safe answer}\EndIf
  \State $u \gets \textsc{Fetch}(id, tmp, rev)$ \Comment{the recorded revision is immutable and fetchable by hash}
  \State \Return $\mathrm{files}(u) = \mathrm{files}(dest)$ byte for byte
\EndFunction
\end{algorithmic}
\end{algorithm}

The decision needs no digests, no staging directories, and no state beyond \texttt{.source}'s two lines, because git's content addressing already provides the merge base: ``unedited'' is verified against upstream \emph{at the recorded revision}, not against a stored hash that could go stale or be forged. The asymmetry is deliberate: a copy that cannot be verified is treated as edited, which routes through the conflict path and can never silently overwrite work.

\subsection{Residency: the auto-trigger set}

\begin{algorithm}[h]
\caption{\textsc{Residency}$()$ --- what fires on its own, as one prompt block}
\label{alg:residency}
\begin{algorithmic}[1]
\State $E \gets$ parse \texttt{.atskills/.autotrigger} \Comment{gitignore syntax; comments dropped; duplicates load once}
\State $R \gets \emptyset$ \Comment{deduplicated by skill directory: one skill loads once}
\State $P \gets$ one ignore-ruleset from all plain lines \Comment{so \texttt{!} negation composes as in git}
\ForAll{$s \in \mathrm{walk}(\texttt{.atskills/})$} \textbf{if} $P$ matches $s.rel$ \textbf{then} $R \gets R \cup \{s\}$
\EndFor
\ForAll{cloud lines $\ell = (@id,\ \mathit{wholedir})$}
  \If{\texttt{.atskills/}$\mathrm{disk}(id)$ holds skills} \Comment{a saved copy answers its own \texttt{@} line}
    \State $R \gets R \cup \mathrm{walk}(\texttt{.atskills/}\Vert\mathrm{disk}(id))$
  \Else
    \State $res \gets \textsc{ReadThroughCache}(id)$ \Comment{once per session; offline serves cache, marked stale}
    \If{$res$ is a skill} $R \gets R \cup \{res\}$
    \ElsIf{$res$ is a menu \textbf{and} $\mathit{wholedir}$} $R \gets R \cup res$
    \Else{} report the line once; \textbf{continue} \Comment{per-line isolation: no line breaks the session}
    \EndIf
  \EndIf
\EndFor
\State \Return block: one row ``\texttt{- name: description (path)}'' per $R$, frontmatter only \Comment{bodies load on trigger}
\end{algorithmic}
\end{algorithm}

The output of \cref{alg:residency} is a plain string, and that is an architectural statement: the client that implements the protocol computes the block, and the host that assembles the model's prompt splices it in verbatim. In AdaL, the TypeScript frontend runs \cref{alg:resolve,alg:residency} (with the shared machine-wide cache at \texttt{\textasciitilde/.cache/atskills}) and the Python backend contains \emph{zero} protocol logic---it stores one string. Any agent whose prompt assembly can accept a string can adopt the protocol without touching its serving layer, and the resolution shown in the management surface can never disagree with what the model receives, because both are the same computation.

\subsection{The checkbox toggle}

\begin{algorithm}[h]
\caption{\textsc{Toggle}$(t)$ --- the \texttt{/skills} checkbox writes the same lines a hand edit would}
\label{alg:toggle}
\begin{algorithmic}[1]
\If{$t$ is a followed cloud row} add or remove its \texttt{@} line \Comment{install $=$ the line, exactly}
\ElsIf{$t$ is checked by its own line} remove that line
\ElsIf{$t$ is covered by a directory line $D/$} \Comment{uncheck under a cover $\Rightarrow$ \textbf{split}}
  \State remove $D/$; add explicit lines for every \emph{other} skill under $D$ that stays on
  \State \Comment{sub-directories that remain whole stay as directory lines --- the file reads true}
\ElsIf{$t$ is covered by a glob pattern} \textbf{refuse}, naming the pattern line to edit \Comment{a pattern has no split}
\Else{} add $t$'s line; a directory row adds $rel/$ and removes now-covered descendant lines
\EndIf
\State \Return a one-line note stating exactly which lines were written or removed
\end{algorithmic}
\end{algorithm}

The toggle is the protocol's answer to ``management for people who will never hand-edit a dotfile'': every checkbox action is defined \emph{as} an edit to \texttt{.autotrigger}, so the dialog, the typed \texttt{/skills} verbs, the \texttt{@} suffixes, and a text editor are four interchangeable ways to produce the same one-line diffs, all reviewed the same way in a pull request. The split rule is the only nontrivial case, and it preserves the file's honesty: after unchecking one skill under a covering directory line, the file names exactly the set that remains on---never a directory line that silently over-claims.

\section{Discussion}
\label{app:discussion}

Four implications follow from the analysis: what \texttt{AGENTS.md}'s dominance teaches, what unification buys, how author incentives change, and what the security posture is.

\minisection{Why \texttt{AGENTS.md} won, and what that implies.} \Cref{app:failures} diagnosed the defection mechanically. The lesson is strategic. Skills lost on \emph{delivery simplicity}, not on content quality, and any successor has to win on the same axis. We cannot tell teams that defected \emph{from} skills apart from teams whose \texttt{AGENTS.md} practice simply came first, but either way delivery decided it. The three-tier model competes on that axis while fixing the costs. Tier 2 is \texttt{AGENTS.md} factored into pieces: the working set lives in the repository just as \texttt{AGENTS.md} does, but each playbook loads on demand at the best position for compliance instead of every instruction taxing every message.

\minisection{Unifying a fragmented integration layer.} One format is scattered over 54 project-level and 58 user-level directories across 75 agents, and even major providers pay per-agent packaging costs (\cref{app:landscape}). The protocol unifies that layer mostly by removing the need to place anything. Referenced skills live nowhere, saved and installed skills live in one agent-neutral project home with one trigger file, and personal collections have a home in the hub. Vendor and home directories keep working as read paths. The result is one resolution model over the same files, an adoption path any agent can join at either level (\cref{sec:ladder}), and existing skills unchanged, so the per-vendor install step becomes optional rather than one more competing island.

\minisection{Author incentives: plays versus album sales.} Install counts measure the wrong thing. An install is a one-time event that says nothing about use, and under install-only delivery a skill outside each user's small installed set is never installed at all, so the long tail reads as zero and authors stop publishing. References are plays rather than album sales. Every load is an act of use, countable per skill, and every published skill has a real chance of being played. Usage-based counting closes the incentive loop that install counts break, for the same reason streaming changed music: distribution stops being gated on shelf space.

\minisection{Alternative: retrieval-based triggering.} A harness could keep all descriptions outside the prompt, embed them, retrieve the top matches for each message, and inject those at the end of the context. That would give auto-triggering without residency, and it would also close tier 1's discovery gap, since a user cannot reference a skill they have never heard of. Production agents already do something similar for tool schemas, loading deferred tools by search instead of keeping them resident. We see this as complementary rather than competing. Retrieval restores implicit discovery for the long tail at the price of bringing back probabilistic activation, while explicit reference keeps loading deterministic. The finder skill and \texttt{:index} are our current answers, and a retrieval layer over the catalog is natural future work.

\minisection{Security considerations.} A referenced skill is instructions fetched from the network and then followed by an agent. That is remote content executed as behavior, the channel indirect prompt injection weaponizes \citep{willison2022promptinjection,greshake2023indirect}. Systematic benchmarking finds no current defense reliably stops injection at the model level \citep{liu2024injectionbench}, and audits of installable LLM-extension ecosystems show they ship exploitable artifacts in practice, with 5.5\% of open-source MCP servers carrying tool-poisoning vulnerabilities \citep{hasan2025mcpfirstglance,invariantlabs2025toolpoisoning,iqbal2024plugins,hou2025mcp}. Provenance and curation, not model-level filtering, have to carry the security burden.

The protocol's properties help more than they hurt. Explicit invocation means the user knows which skill loaded and when, whereas auto-triggered skills load silently. Because content arrives as visible tool output, the full source, including every reference file, can be inspected in the transcript instead of executing unseen from a config directory. References the agent makes on its own bring implicit activation back and must appear in the transcript exactly as user references do (\cref{fig:hero}). Tier 1 leaves no installed behavior behind, though side effects of any script it runs are bounded by the session's tool permissions, as with any agent action. On-demand delivery also allows a vetting step that install-first delivery structurally lacks. Because a reference is a single-session act, an unfamiliar skill can be trialed in a sandbox, whether an isolated worktree, a container, or a restricted-permission session, and watched end to end before it is granted any persistence. Installation instead places a skill into every future session sight unseen. Saving puts the full skill text in a git diff, open to ordinary code review, which holds the team to a reviewed copy rather than a mutable upstream, and the \texttt{.source} stamp records the exact upstream revision taken, so ``did we change it or did they'' is always answerable. Tier 3 activation runs through the same channel, since adding a line to \texttt{.autotrigger} is a one-line diff, so a change to the team's resident prompt is reviewed before it reaches anyone's session; script execution from followed skills is confirmed by revision change, never silently (\cref{sec:autotrigger}). Catalog slugs inherit the usual package-namespace risks such as typosquatting and name confusion, and digests and provenance metadata are the mitigations. Injection risk from malicious skill content is real but not new. It is the same for installed skills, except that an installed skill's body loads invisibly.

\minisection{Limitations and disclosure.} This is a design paper grounded in ecosystem measurement, published-skill evidence, and the positional-attention literature. We have not yet run controlled experiments isolating trigger reliability as a function of install count, or compliance as a function of injection position, for skill-sized instructions specifically. The token measurements come from one real working setup rather than a survey. If future architectures or harness-side calibration reduce positional bias, the residency argument weakens, though the standing-tax, dilution, and lifecycle arguments do not. A referenced skill also drifts from the task in very long sessions, and re-referencing is the cheap, deterministic remedy. Tier 1 puts a network round trip on the critical path, though the validating cache reduces repeated use to one revision probe and serves the cached copy offline; a reference still resolves to living upstream content by design, so for anything the team must control the answer is a save---a local, reviewed copy---rather than a version pin the protocol deliberately does not offer. Finally, a disclosure. The author founded SylphAI, which develops AdaL and operates the catalog and hub described here, so this paper advocates infrastructure its author runs. The format and protocol are open precisely so that neither is required: \texttt{@skills:gh:} references and plain self-hosting work with no catalog at all.

\section{Related Work}
\label{app:related}

\minisection{Positional attention and long-context limits.} \citet{liu2024lost} established that language models attend most reliably to the beginning and end of long contexts, with systematic degradation in the middle. Subsequent work strengthened the finding: the bias is an intrinsic, U-shaped attention artifact present regardless of content relevance \citep{hsieh2024found}, mirrors human serial-position (primacy/recency) effects and resists prompt-level mitigation \citep{guo2024serial}, and effective context is far smaller than advertised context across model families \citep{hsieh2024ruler}. Input length alone---holding the task fixed---degrades reasoning well below nominal limits \citep{levy2024same}, and irrelevant context measurably distracts models even when they are told to ignore it \citep{shi2023distracted}. This literature supplies the mechanical basis for our delivery claims (\cref{sec:failure-position,sec:failure-tax}).

\minisection{Instruction density and multi-turn decay.} Complementary evidence tracks what happens to \emph{instructions} specifically: adherence falls as the number of simultaneous instructions grows, to 68\% for the best frontier model at 500 constraints \citep{jaroslawicz2025instructions}; system-message compliance decays over conversation turns \citep{qin2024sysbench}; and models lose an average of 39\% in long multi-turn settings versus single-turn on identical tasks \citep{laban2025lost}. In-context learning is similarly sensitive to what is in the prompt and where---example selection and ordering alone swing performance between near state-of-the-art and near chance \citep{lu2022ordering,zhao2021calibrate}. Together these justify treating resident-instruction capacity as the scarce budget of \cref{sec:failure-structural}; prompt-compression work \citep{jiang2023llmlingua} attacks the same cost from the other side.

\minisection{Retrieval-augmented generation.} RAG \citep{lewis2020rag,gao2023ragsurvey} retrieves \emph{declarative} knowledge at inference time rather than storing it in weights; \citet{rubin2022retrieveprompts} extended retrieval to prompt content itself, and Self-RAG made the retrieval decision adaptive---fetch only when needed \citep{asai2024selfrag}. Tier-1 reference is the analogous move for \emph{procedural} knowledge: rather than storing instructions in the prompt (the ``weights'' of a session), fetch them at the moment of need. The difference is the trigger: RAG retrieval is implicit and similarity-based; \atskill{} references are explicit and deterministic, which is what reliability requires for instructions as opposed to facts.

\minisection{Agent memory and skill libraries.} \citet{weng2023agent} framed the canonical agent as planning, memory, and tools, with the context window as bounded working memory. Agents that accumulate reusable text-form knowledge validate the premise that procedural text substitutes for weight updates: Voyager grows a library of executable skills and transfers it to new worlds \citep{wang2024voyager}; Reflexion stores verbal reflections that improve later trials \citep{shinn2023reflexion}; ExpeL distills cross-task insights recalled at inference \citep{zhao2024expel}. Closest to our design, Agent Workflow Memory induces reusable workflows from trajectories and \emph{selectively provides} them to the agent, improving web-task success by 24--51\% relative \citep{wang2025awm}---retrieval-not-residency, demonstrated. All of these are \emph{self-authored, per-agent} accumulations; the \atskill{} catalog externalizes the same loop into a shared, human-curated commons with distribution.

\minisection{Tool learning at scale and agent protocols.} Toolformer \citep{schick2023toolformer} and ReAct \citep{yao2023react} established models invoking external capabilities; at ecosystem scale, retrieval becomes architecturally mandatory---Gorilla pairs the model with a documentation retriever precisely because baked-in API knowledge goes stale \citep{patil2024gorilla}, and ToolLLM equips its agent with a neural API retriever because 16{,}464 APIs cannot fit in context \citep{qin2024toolllm}. The same argument at 56{,}804 skills is this paper. The Model Context Protocol \citep{anthropic2024mcp,hou2025mcp} standardizes agent-to-service transport; MCP unifies how agents call \emph{services}, \atskill{} unifies how agents load \emph{instructions}, and the two compose (\cref{sec:zero-integration}). Production harnesses likewise load deferred tool schemas by search rather than keeping them resident, the same residency-avoidance move applied to tools instead of instructions. Coding agents---the deployment surface for skills---operate in repo-scale settings where surrounding structure, not raw model capability, drives performance \citep{jimenez2024swebench,yang2024sweagent}.

\minisection{The skills ecosystem in practice.} Anthropic's Agent Skills define the format this protocol delivers \citep{anthropic2025skills,anthropic2025skillseng}; skills.sh \citep{vercel2025skills} indexes the public corpus, popularized one-command installation, and has since added per-skill selection and a \texttt{use} command that emits one skill as a prompt without installing it---the closest neighbour to this work, and an independent signal that on-demand delivery is where the ecosystem is heading. It differs in what we argue is the load-bearing part: a skill is selected by name inside a repository rather than addressed by path, exactly one skill resolves per invocation (a collection or subtree has no expressible form), and nothing is cached between calls, so use re-fetches while the install lifecycle it sits beside---lockfiles, per-agent directories, update and remove---remains intact underneath; the discovery convention adds self-hosted publishing with digests \citep{agentskills2025discovery}; and \texttt{AGENTS.md} \citep{agentsmd2025} is the competing convention whose dominance \cref{app:discussion} analyzes. Practitioner analysis converges on our premises from experience rather than measurement: \citet{willison2025skills} attributes skills' promise to on-demand loading and token frugality against MCP's tens-of-thousands-of-tokens residency; \citet{osmani2026skills} demonstrates one \texttt{SKILL.md} corpus portable across five harnesses; \citet{anthropic2025context} prescribes just-in-time retrieval and treating context as a finite attention budget; and \citet{karpathy2025software} frames the context window as the RAM of an LLM operating system; MemGPT built exactly that virtual-memory design, paging content between context and external storage \citep{packer2023memgpt}. In that framing, install-only skills are programs pinned permanently in RAM, and \atskill{} is paging.

\minisection{The \texttt{@} convention.} Explicit context addressing via \texttt{@}-mentions is the de facto interaction standard across coding agents. AdaL \citep{adal2026} and Claude Code mention both individual files and whole directories with \texttt{@path}; Cursor's \texttt{@}-symbols scope files, folders, documentation, and web results \citep{cursor2025atsymbols}; GitHub Copilot exposes \texttt{@workspace} \citep{github2024copilot}; and the same affordance---files, folders, symbols, URLs, terminal state---is documented in Windsurf, Continue, Cline, Sourcegraph Cody, Zed, Amazon Q, Gemini CLI, and JetBrains AI Assistant \citep{cline2026docs,continue2026docs,sourcegraph2026docs}. Across a dozen agents, \texttt{@} is how users deliberately place something into the context window---yet every one of these mechanisms addresses \emph{data}; none addresses \emph{procedures}. \atskill{} completes the pattern: the same gesture, applied to the one resource type that today can only be installed.

\minisection{Package management.} The install-centric skills lifecycle imports the package-manager mental model (registry, install, update), but software packages are the wrong analogy for prompt content: code costs nothing until called, while a resident description costs attention on every message. The closer analogy is the open-format-plus-hub pattern---git and GitHub, model weights and Hugging Face---where an open artifact enabled a hosted service to make management effortless, and the value accrued to distribution and curation rather than format ownership (\cref{sec:hub}).

\end{document}